\documentclass[10pt,twocolumn,letterpaper]{article}

\usepackage{cvpr}              %
\usepackage{subfiles}


\usepackage[table,xcdraw]{xcolor}
\usepackage{array}

\def\thickhline{\noalign{\hrule height 1pt}}
\newcolumntype{P}[1]{>{\centering\arraybackslash}p{#1}}
\newcolumntype{M}[1]{>{\centering\arraybackslash}m{#1}}

\newcommand{\mparagraph}[1]{\vspace{1mm}\noindent{\textbf{#1.}\hspace{1mm}}}

\usepackage{import} %
\graphicspath{{figs/}}

\usepackage[ruled,noend]{algorithm2e} %

\SetAlFnt{\small}
\SetAlCapFnt{\small}
\SetAlCapNameFnt{\small}
\SetAlCapHSkip{0pt}
\IncMargin{-\parindent}
\usepackage{ifpdf}
\usepackage{graphicx}
\ifpdf
  
\else
\fi
\usepackage{color}
\definecolor{cyan}{cmyk}{1,0,0,0}
\definecolor{darkgreen}{rgb}{0,0.5,0}
\definecolor{orange}{rgb}{1,0.5,0}
\definecolor{magenta}{cmyk}{0,1,0,0}
\definecolor{darkyellow}{cmyk}{0,0,0.75,0}
\definecolor{gray}{rgb}{0.8,0.8,0.8}

\usepackage{amsmath} %

\usepackage[toc,page]{appendix} %
\usepackage{wrapfig} %
\usepackage{comment}
\usepackage{bm}
\usepackage{cuted} %
\usepackage{lipsum} %
\usepackage{mathtools} %
\usepackage{algorithmicx}
\usepackage{algpseudocode}
\usepackage{nicefrac}

\usepackage{gensymb}
\usepackage{float}
\usepackage{soul}
\usepackage{array}
\usepackage{multirow}
\usepackage{hhline}
\usepackage{setspace}
\usepackage{nicefrac}

\algdef{SE}[DOWHILE]{Do}{doWhile}{\algorithmicdo}[1]{\algorithmicwhile\ #1}%

\makeatletter
\renewcommand{\ALG@beginalgorithmic}{\small}
\makeatother

\newcommand{\DELETE}[1]{} %
\newcommand{\IGNORE}[1]{}
\usepackage{datenumber}
\usepackage{calc}
\usepackage[mmddyyyy]{datetime}

\newcounter{datetoday}
\newcounter{diffyears}
\newcounter{diffmonths}
\newcounter{diffdays}

\newcommand{\difftoday}[3]{%
      \setmydatenumber{datetoday}{\the\year}{\the\month}{\the\day}%
      \setmydatenumber{diffdays}{#1}{#2}{#3}%
      \addtocounter{diffdays}{-\thedatetoday}%
      \ifnum\value{diffdays}>0
        \def\diffbefore{}%
        \def\diffafter{left}%
      \else
        \def\diffbefore{}%
        \def\diffafter{ago}%
        \setcounter{diffdays}{-\value{diffdays}}%
      \fi
      \setcounter{diffyears}{\value{diffdays}/365}%
      \setcounter{diffdays}{\value{diffdays}-365*\value{diffyears}}%
      \setcounter{diffmonths}{\value{diffdays}/30}%
      \setcounter{diffdays}{\value{diffdays}-30*\value{diffmonths}}%
      \diffbefore
      \ifnum\value{diffyears}=0
      \else
        \ifnum\value{diffyears}>1
            \thediffyears\space years,
        \else
            \thediffyears\space year,
        \fi
      \fi
      \ifnum\value{diffmonths}=0
      \else
        \ifnum\value{diffmonths}>1
            \thediffmonths\space months
        \else
            \thediffmonths\space month
        \fi
      \fi
      \ifnum\value{diffdays}=0
      \else
        \ifnum\value{diffdays}>1
            \thediffdays\space days
        \else
            \thediffdays\space day
        \fi
      \fi
      \diffafter
}

\definecolor{cvprblue}{rgb}{0.21,0.49,0.74}
\usepackage[pagebackref,breaklinks,colorlinks,citecolor=cvprblue]{hyperref}
\usepackage[accsupp]{axessibility}
\def\paperID{3021} %
\def\confName{CVPR}
\def\confYear{2024}

\title{OmniLocalRF: Omnidirectional Local Radiance Fields from Dynamic Videos}

\author{
Dongyoung Choi
\hspace{15mm}
Hyeonjoong Jang
\hspace{15mm}
Min H. Kim
\\[2mm]
KAIST
}

\begin{document}

\maketitle

\begin{abstract}
\noindent 
Omnidirectional cameras are extensively used in various applications to provide a wide field of vision. However, they face a challenge in synthesizing novel views due to the inevitable presence of dynamic objects, including the photographer, in their wide field of view. In this paper, we introduce a new approach called Omnidirectional Local Radiance Fields (OmniLocalRF) that can render static-only scene views, removing and inpainting dynamic objects simultaneously. Our approach combines the principles of local radiance fields with the bidirectional optimization of omnidirectional rays. Our input is an omnidirectional video, and we evaluate the mutual observations of the entire angle between the previous and current frames. To reduce ghosting artifacts of dynamic objects and inpaint occlusions, we devise a multi-resolution motion mask prediction module. Unlike existing methods that primarily separate dynamic components through the temporal domain, our method uses multi-resolution neural feature planes for precise segmentation, which is more suitable for long 360{\degree} videos. Our experiments validate that OmniLocalRF outperforms existing methods in both qualitative and quantitative metrics, especially in scenarios with complex real-world scenes. In particular, our approach eliminates the need for manual interaction, such as drawing motion masks by hand and additional pose estimation, making it a highly effective and efficient solution.
\end{abstract}
    
\section{Introduction}
\label{sec:intro}
Omnidirectional cameras such as Ricoh Theta or Insta360 allow capturing panoramic 360{\degree} views in a single shot.
Various applications with omnidirectional images such as spherical depth estimation~\cite{zioulis2018omnidepth, won2019sweepnet, zio19spherical}, novel view synthesis~\cite{broxton2020immersive, broxton2020deepview, broxton19lowcost, overbeck18system, pozo19integrated, OmniPhotos, Choi_2023_CVPR, Egocentric, barron2022mipnerf360} and geometry reconstruction~\cite{Egocentric, OmniPhotos} aiming at large-scale static scenes have recently been explored. In particular, synthesizing 360{\degree} novel views can provide continuous views from unobserved camera angles while maintaining its details.

\begin{figure}
	\centering
	\vspace{-6mm}
	\includegraphics[width=1.0\linewidth]{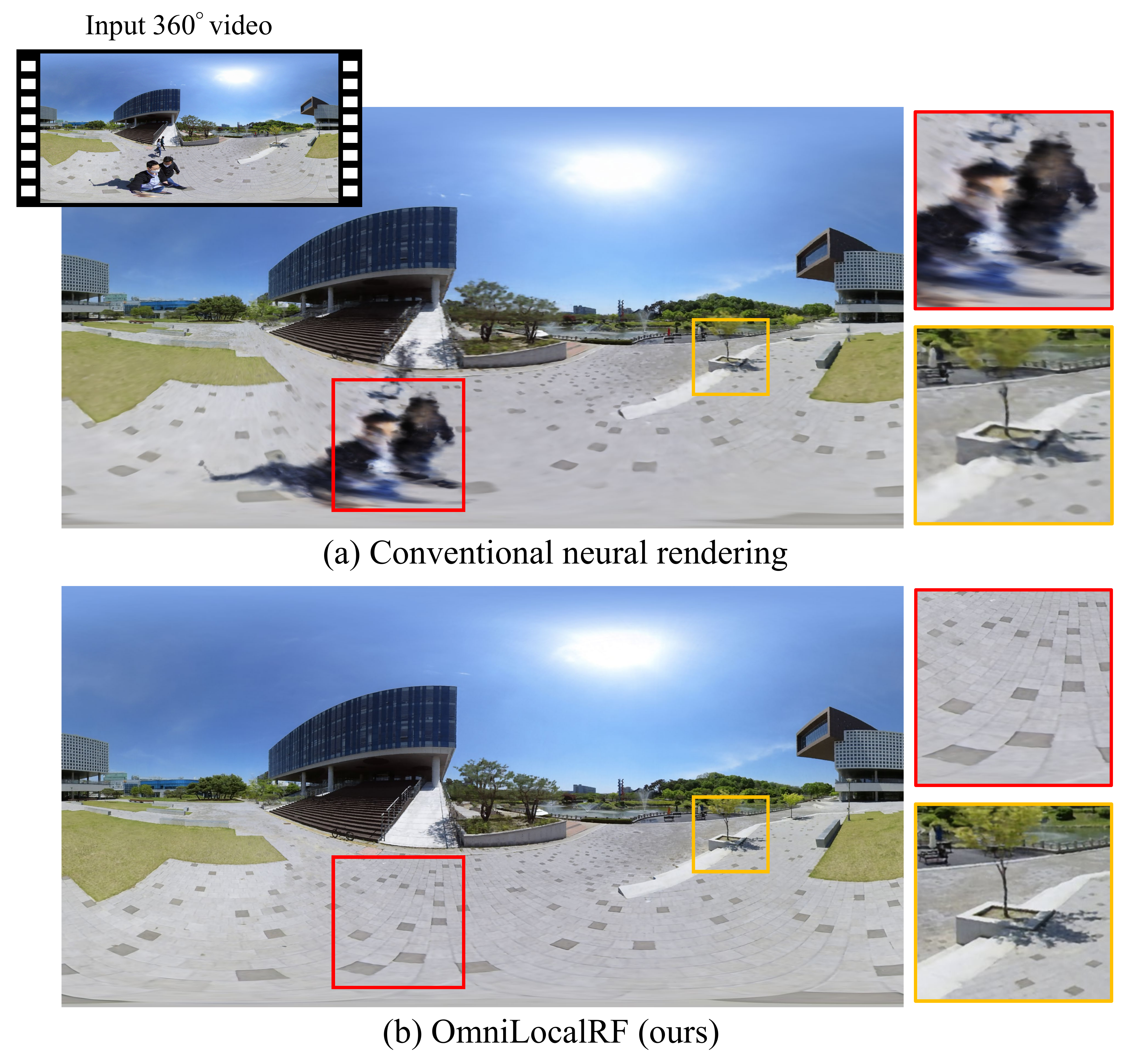} 
	\vspace{-8mm}
	\caption{\label{fig:teaser}
	We introduce omnidirectional local radiance fields for photorealistic view synthesis of static scenery from 360{\degree} videos.
Our method effectively removes dynamic objects (including the photographer) without manual interaction. Also, it achieves high-resolution details in the inpainted regions by means of bidirectional observations of omnidirectional local radiance fields.
Refer to the supplemental video for more results.}
	\vspace{-6mm}
\end{figure}

However, recent novel view synthesis methods struggle to apply to omnidirectional input for the following reasons.
When capturing omnidirectional videos to record static environments, dynamic objects are prone to be captured as an extension of the field of view, and capturing a photographer is inevitable unless employing a dedicated hardware or remote controller.
When synthesizing novel views, these captured objects are represented as ghosting artifacts onto the rendered results~\cite{sabour2023robustnerf}. Despite these problems, existing methods have achieved 360{\degree} view synthesis, relying on constrained capturing conditions, where it minimizes the advent of dynamic objects~\cite{OmniPhotos, Choi_2023_CVPR} or requiring dedicated hardware~\cite{broxton2020immersive, broxton2020deepview, broxton19lowcost, overbeck18system, pozo19integrated}, which are not suitable for casual 360{\degree} photography.

Low-rank decomposition through robust principal component analysis~\cite{zhou2012moving, guyon2012foreground, guyon2012moving} and existing optimization-based methods~\cite{granados2008background, granados2012not} effectively eliminate dynamic objects on the image domain. However, their applicability is limited to scenarios involving multiple images captured from the same viewpoints. 
Recent view synthesis works~\cite{tschernezki21neuraldiff, wu2022d2nerf, park2021hypernerf, park2021nerfies} detach dynamic objects during static view synthesis by modeling dynamic objects along the temporal domain. However, they are inappropriate for long 360{\degree} videos because of the neural model's capacities.

In this paper, we propose omnidirectional local radiance fields (OmniLocalRF) that can render novel views of static scene environments from casual dynamic 360{\degree} videos. 
We formulate local radiance fields (LocalRF)~\cite{meuleman2023localrf} into a novel bidirectional training scheme designed explicitly for omnidirectional video input.
We develop a module that uses multi-resolution neural feature planes to predict motion masks of every frame and segment frame-dependent components.
Our LocalRF-based approach enables us to synthesize views and estimate camera poses from long videos.
We automatically remove and inpaint dynamic objects while performing large-scale omnidirectional view synthesis. See Figure~\ref{fig:teaser}.

It is worth noting that different from conventional perspective cameras, omnidirectional cameras capture continuous scene information across multiple frames. This unique characteristic allows us to design a novel optimization approach that involves the bidirectional evaluation of samples taken from distant frames through an omnidirectional contraction scheme. This method enables us to effectively remove and inpaint broken dynamic objects through backward refinement, demonstrating omnidirectional reconstruction and decoupling of dynamic objects across complex real-world scenes.
In summary, our contributions are:
\begin{itemize}[leftmargin=0.8cm] 
	\item A view synthesis method based on bidirectional optimization through distant frames while conserving locality across radiance fields,
	\item A new motion mask prediction module that accurately segments dynamic objects, even in 360{\degree} videos without requiring a pretrained model, and
	\item A camera pose estimation technique based on local view synthesis of 360{\degree} videos.
\end{itemize}
Our code is freely available for research purposes\footnote{\url{https://vclab.kaist.ac.kr/cvpr2024p1/}}.

\section{Related work}
\label{sec:relatedwork}
\mparagraph{Omnidirectional view synthesis}
OmniPhotos~\cite{OmniPhotos} and Jang et al.~\cite{Egocentric} reconstruct geometry from 360{\degree} videos using mesh representation. Several approaches~\cite{broxton2020immersive, broxton2020deepview, broxton19lowcost, overbeck18system, pozo19integrated} use a spherical camera rig to render images at unobserved camera positions. MatryODShka~\cite{Attal20ECCV} proposes a multi-sphere image representation for omnidirectional view synthesis. However, these methods do not work for 360{\degree} videos casually taken with dynamic objects.

Recent advancements for rendering 360{\degree} images, e.g., Mip-NeRF360~\cite{barron2022mipnerf360} and EgoNeRF~\cite{Choi_2023_CVPR}, use volume rendering to create omnidirectional images. Mip-NeRF360 extends the capabilities of Neural Radiance Fields (NeRF)~\cite{mildenhall2020nerf} to unbounded scenes, while EgoNeRF uses the Yin-Yang grid to obtain a balanced polar representation of neural features. However, these methods limit use in scenarios with dynamic objects in the training dataset, as they are designed for egocentric scenes captured using a selfie stick or by excluding dynamic objects with a motion mask.

\mparagraph{Removing dynamic objects}
Novel view synthesis methods generally aim to reconstruct static objects through the multi-view stereo~\cite{mildenhall2020nerf, barron2022mipnerf360, tancik2022blocknerf, Turki_2022_CVPR}. In existing works, separating dynamic and static objects by pretrained model~\cite{he2017mask, chen2018encoder} on input images and modeling them respectively enables rendering static geometry if multi-view of dynamic objects are sufficiently provided in a dataset~\cite{park2021hypernerf, park2021nerfies, liu2023robust, li2021neural}. 
In 360{\degree} videos that do not provide enough multi-view cues of dynamic objects, modeling them across the 3D space is challenging due to geometric inconsistency over time. 

To reduce ghosting artifacts caused by dynamic objects in view synthesis, several models \cite{rematas2022urban,tancik2022blocknerf} exclude dynamic objects using an external segmentation model \cite{cheng2020panoptic, chen2018encoder}. However, these models fail to mask out objects that are not labeled, such as shadows. Recent works, such as D$^2$NeRF~\cite{wu2022d2nerf} and Neuraldiff~\cite{tschernezki21neuraldiff}, segment dynamic objects by reconstructing them in 3D space across the temporal domain, but require an extensive parameterization, making them less scalable for long 360{\degree}  videos. OmnimatteRF~\cite{Lin_2023_ICCV} estimates delicate motion masks from rough masks provided by Mask R-CNN~\cite{he2017mask} with optical flow~\cite{teed2020raft}, but still needs a pretrained segmentation model. RobustNeRF~\cite{sabour2023robustnerf} discriminates between inliers and outliers based on photometric error and down-weights the outliers to decrease the effect of dynamic objects during training. Nevertheless, using a limited number of ray samples driven by down-weighting outliers slows down the training procedure, and these models are not scalable for long 360{\degree} videos.

\mparagraph{Pose estimation}
Conventional view synthesis relies on Structure from Motion (SfM) methods, like COLMAP~\cite{schoenberger2016sfm} and OpenMVG~\cite{moulon2017openmvg}, for camera poses, but these can be challenging to compute for large-scale data, leading to poor view synthesis quality. Several NeRF variants \cite{jeong2021self,wang2021nerf,lin2021barf,bian2023nope} optimize radiance fields and poses jointly, but struggle to estimate camera poses for 360{\degree} videos with dynamic objects \cite{meuleman2023localrf}. RoDynRF~\cite{liu2023robust} uses Mask R-CNN to separate dynamic and static components, but it is not scalable and requires a pretrained segmentation model. LocalRF~\cite{meuleman2023localrf} succeeds in long trajectory pose calibration but is susceptible to artifacts caused by dynamic objects.

\section{Omnidirectional Local Radiance Fields}
\label{sec:method}
\mparagraph{Overview}
Our goal is to generate photorealistic static scenes from unobserved viewpoints using long 360{\degree} videos, including dynamic objects as input. 
We optimize multiple radiance fields with continuous local frames while sliding frame windows and produce high-quality view synthesis.
However, training blocks solely with local frames can result in ghosting artifacts from dynamic objects. 
To address this, we use an omnidirectional local radiance fields approach and a motion mask module to separate dynamic and static objects.
We also propose a novel bidirectional optimization method to enhance the stability of static structural representations and eliminate residual artifacts. Our method produces superior rendering results than existing methods.

\subsection{Preliminaries}\label{subsec:preliminaries}
We render the color $\hat{\mathbf{C}}(\mathbf{r})$ by volume rendering samples $\mathbf{x}_i$ whose density and color are $(\sigma_i,\mathbf{c}_i)$ along a ray $\mathbf{r}$ and train the radiance fields $\bold{RF}_{\Theta}$ to predict $(\sigma_i,\bold{c}_i)$ from the L1 photometric error between $\hat{\mathbf{C}}(\mathbf{r})$ and input image $\mathbf C(\mathbf r)$
as it is more robust against outliers than MSE:
\begin{equation}
	\mathcal{L}_\text{rgb} =\left\| \hat{\mathbf{C}}(\mathbf{r}) - \mathbf {C( \mathbf r)} \right\|_1.
\end{equation}
To extend NeRF to cover a large scale, we also use the contraction equation \cite{barron2022mipnerf360} over sample points:
\begin{equation}
	\label{eq:forward_rgb}
	\resizebox{0.8\linewidth}{!}{
	\mbox{\fontsize{10}{12}\selectfont $
\text{contract}(\mathbf{x}) = \left\{ {\begin{array}{*{20}{c}}
		\mathbf{x} & {\left\| \mathbf{x} \right\| \le 1} \\
		{\left( {2 - \frac{1}{\left\| \mathbf{x} \right\|}} \right)\left( {\frac{\mathbf{x}}{{\left\| \mathbf{x} \right\|}}} \right)} & 	{\left\| \mathbf{x} \right\| > 1}
\end{array}} \right.,
	$ } } %
\end{equation}
Using the contraction function in radiance fields, which maps world coordinates onto a contracted space, aids in large-scale view synthesis by focusing nearby regions while representing distant components~\cite{tancik2023nerfstudio, meuleman2023localrf}.
However, they face challenges in long camera trajectories due to static model allocation.

Our approach uses multiple TensoRFs~\cite{chen2022tensorf} to perform view synthesis and camera registration from videos, following LocalRF~\cite{meuleman2023localrf}. 
We allocate a NeRF block $\mathbf{RF}_{\Theta _{m}}$, where $m \in \{1,\cdots,M\}$, and insert input frames $\mathbf{C}_k$ with the corresponding camera poses ${[R\mid t]_k}$, $k \in \{1,\cdots,K\}$ into a temporal window $\mathcal{W}_m=\left\{ (\mathbf{C}_{w(m,1)}, [R\mid t]_{w(m,1)}),\cdots, (\mathbf{C}_{w(m,N)},[R\mid t]_{w(m,N)}) \right\}$, where $w(m,n)$ denotes the frame number of $m$-th windows' $n$-th frame. 
If the distance between the camera $[R\mid t]_{w(m,N)}$ and the block center becomes too large, we stop inserting frames and optimize the block $\mathbf{RF}_{\Theta _{m}}$ with the camera poses $[R\mid t]_{w(m,n)}$ where $n \in \{1,\cdots,N\}$ using $\mathcal{W}_m$. 
After optimizing a $\mathbf{RF}_{\Theta _{m}}$ block, we create a new block $\mathbf{RF}_{\Theta _{m+1}}$ with its window $\mathcal{W}_{m+1}$ and insert frames until the end of the videos.

\begin{figure}[tp]
	\centering
	\vspace{-3mm}	
	\includegraphics[width=0.45\textwidth]{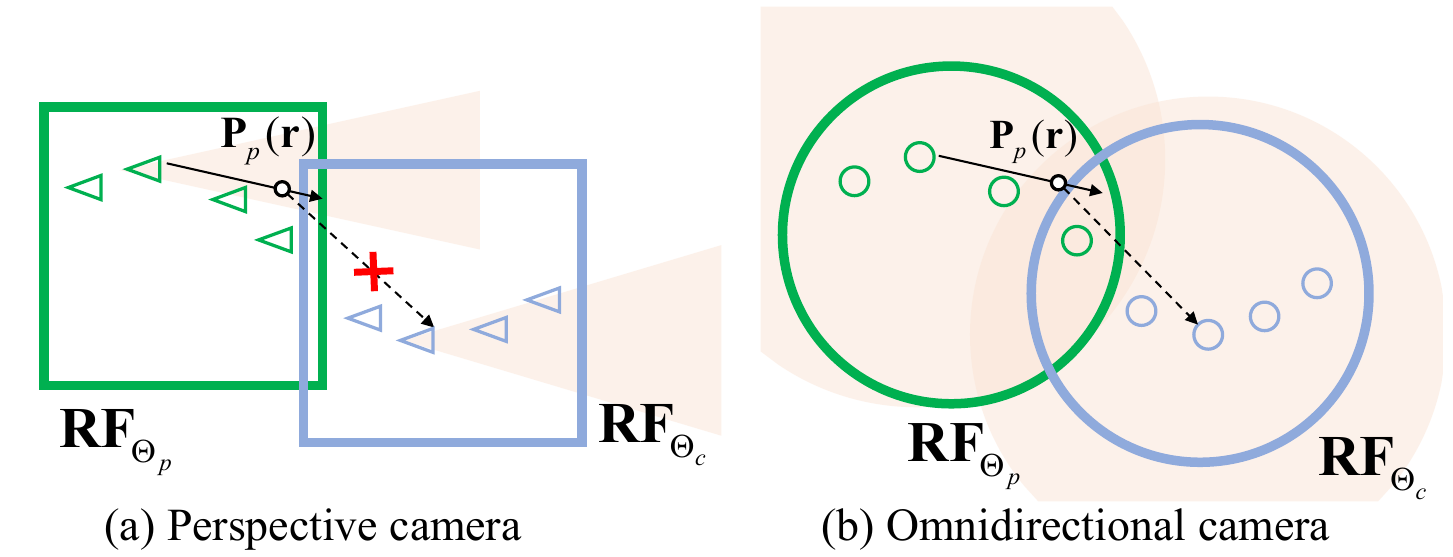} 
	\vspace{-3mm}
	\caption{\label{fig:camera_diff}
	In the perspective video of marching forward, optimized radiance blocks $\mathbf{RF}_{\Theta_p}$ may not be visible in the frame used to train current radiance fields $\mathbf{RF}_{\Theta_c}$. 
	However, in omnidirectional video, every uncontracted space of the optimized blocks can be seen, enabling effective bidirectional optimization. The boundary indicates the radiance fields' focusing region uncontracted.}
	\vspace{-4mm}
\end{figure}

\begin{figure*}[tp]
	\centering
	\vspace{-8mm}
	\includegraphics[width=0.8\textwidth]{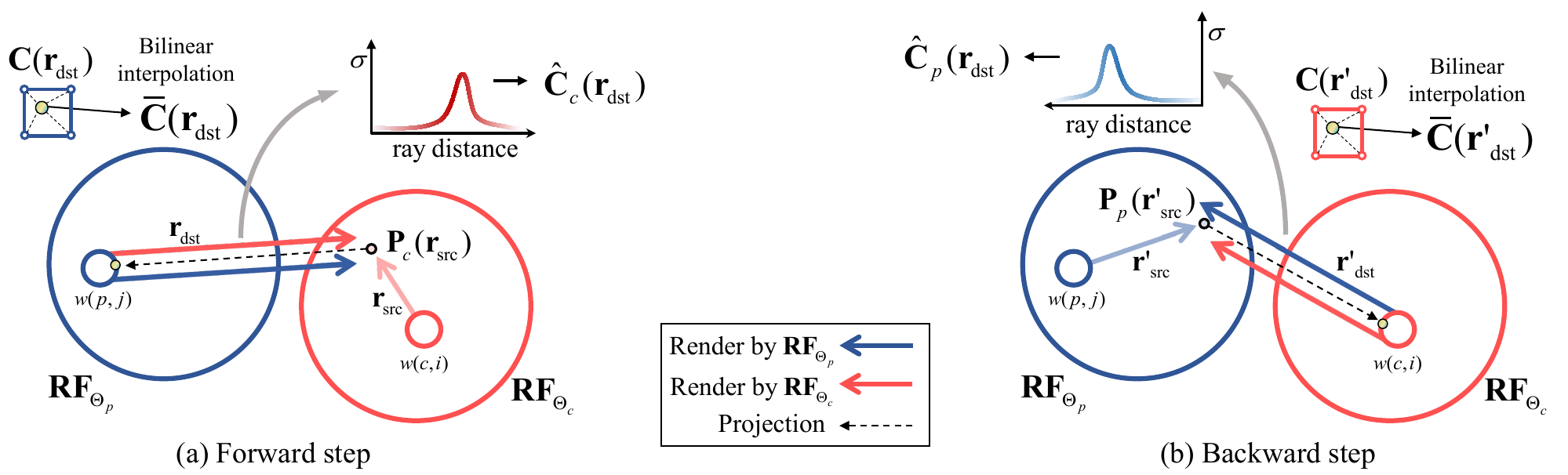}%
	\vspace{-3mm}
	\caption{\label{fig:brdf}
Our bidirectional optimization for omnidirectional videos. 
(a) In the forward step, we project the point $\mathbf{P}_c(\mathbf{r}_\text{src})$ rendered by $\mathbf{RF}_{\Theta_c}$ to the 
destination frame $w(p,j)$, used to train the previous radiance block $\mathbf{RF}_{\Theta_p}$. We then render the color and depth through $\mathbf{RF}_{\Theta_c}$ and $\mathbf{RF}_{\Theta_p}$, respectively, and use the L1 photometric error between the fully rendered color $\hat{\mathbf{C}}_c(\mathbf{r}_\text{dst})$ and the bilinearly interpolated input image $\bar{\mathbf{C}}(\mathbf{r}_\text{dst})$ (Eq.~\eqref{eq:forward_rgb}) 
to update $\mathbf{RF}_{\Theta_c}$ and a mask module.
(b) In the backward step, we switch the source and destination frames and refine $\mathbf{RF}_{\Theta_p}$ through the valid rays from static areas that meet $\mathcal{R}_\text{P}$.}
	\vspace{-4mm}
\end{figure*}

\subsection{Bidirectional Optimization by Distant Frames} \label{subsec:brdf}
Previous reconstruction approaches~\cite{Choi_2023_CVPR, sabour2023robustnerf, barron2022mipnerf360, wang2023f2} predefine the reconstruction range before optimization and train a single NeRF-based module with registered poses. 
LocalRF~\cite{meuleman2023localrf} progressively allocates NeRF modules and performs large-scale reconstruction in a local manner without requiring camera priors.
However, the vanilla LocalRF has limitations in leveraging global information and is susceptible to slowly moving objects that appear stationary from a local perspective. 
To overcome these issues, we propose a novel solution that leverages the omnidirectional nature of 360{\degree} view synthesis while taking advantage of locality. This improves overall reconstruction quality.

As shown in Figure~\ref{fig:camera_diff}, in perspective videos of marching forward, distant frames contain only a small amount of information needed to refine poses and radiance fields. 
This is inefficient and can lead to instability. However, in 360{\degree} videos, distant frames contain a substantial amount of data to refine radiance fields. 
As a result, we propose a bidirectional optimization by distant frames as part of progressive optimization, which can globally refine radiance fields and poses for omnidirectional view synthesis.

\mparagraph{Forward step}
As we progressively optimize multiple radiance fields $\mathbf{RF}_{\Theta_{m}}$, there are $\mathbf{RF}_{\Theta_c}$ that OmniLocalRF currently optimizes and a group of $\mathbf{RF}_{\Theta_p}, p \in \{1,\cdots,c-1 \}$, which are already converged. In LocalRF, $\mathbf{RF}_{\Theta_c}$ is trained only with its temporal window $\mathcal{W}_c$ using the photometric loss defined as:
\begin{equation}
	\mathcal{L}^{\text{for}}_{\text{rgb, s}}=	\sum\limits_{\mathbf{r}_\text{src} \in \mathcal{R}}  \left\| {\hat{\mathbf{C}}_c(\mathbf r_\text{src}) - {\mathbf{C}}(\mathbf r_\text{src})} \right\|_1, \label{eq:main_rgb}
\end{equation}
where $\mathbf r_\text{src}$ represents a ray from the current window $\mathcal{W}_c$, limiting the use of global frames. 
Here we additionally refine $\mathbf{RF}_{\Theta_c}$ using prior frames outside  $\mathcal{W}_c$ and refer to it as a forward step since it proceeds in the same direction as window sliding.

As shown in Figure~\ref{fig:brdf}(a), to concentrate on the areas where $\mathbf{RF}_{\Theta_c}$ occupies, we render a depth value
through $\mathbf{RF}_{\Theta_c}$ at the pixel $\mathbf p_\text{src}$, the origin of $\mathbf r_\text{src}$, on $w(c,i)$-th frame, $w(c,i) \in \mathcal{W}_c$, and obtain projected pixels $\mathbf p_\text{dst}$ to the $w(p,j)$-th frame which is a randomly selected member in $\mathcal{W}_p$ but not included in $\mathcal{W}_c$ by 
\begin{equation}
	\mathbf {p}_\text{dst} = {\Pi} {\left( {{{[R\mid t]}_{w(c,i) \to w(p,j)}}{\Pi} {^{ - 1}\left(\mathbf{p}_\text{src},\hat D_c({\mathbf{r}_\text{src}}) \right)} } \right)}. \label{eq:p_dst}
\end{equation}
Here $\hat D_c$ is a rendered depth value using $\mathbf{RF}_{\Theta_c}$, and ${{[R\mid t]}_{w(c,i) \to w(p,j)}}$ is the relative camera matrix from $w(c,i)$-th to $w(p,j)$-th frame.
${\Pi()}$ denotes the equirectangular projection operator, and ${\Pi}^{-1}()$ backprojects a pixel on an equirectangular image to the world space.

We then render a depth value
and a color value 
along ${\mathbf{r}_{\text{dst}}}$ which casts from  $\mathbf{p}_{\text{dst}}$ through $\mathbf{RF}_{\Theta_{p}}$ and $\mathbf{RF}_{\Theta_c}$, respectively.
In order to exclusively use the reliable samples, we choose valid rays which satisfy
\begin{equation}
	\label{eq:forward_condition}
	\resizebox{1.0\linewidth}{!}{
	\mbox{\fontsize{10}{12}\selectfont $
	\mathcal{R}_\text{C} = \{ 
	\mathbf{r} \, | (1-\mathcal{T})\hat{D}_{p}(\mathbf{r}) \leq \hat{D}_{c}(\mathbf{r}) 
	\leq (1+\mathcal{T})\hat{D}_{p}(\mathbf{r}) ,  \hat{D}_{c}(\mathbf{r}) \leq 1  \},
	$ } } %
\end{equation}
where $\hat{D}_{p}$ is a rendered depth by $\mathbf{RF}_{\Theta_{p}}$, and $\mathcal{T}$ is the valid margin of photometric refinement, constant at $0.05$.
The regions satisfying Eq.~\eqref{eq:forward_condition} are co-visible from two RF blocks. 
In cases where static geometry exists, causing occlusion, the rendering depths between the two RF blocks differ significantly, preventing the execution of bidirectional optimization.
We update $\mathbf{RF}_{\Theta_c}$ and a mask module from the conventional photometric loss in Eq.~\eqref{eq:main_rgb} with 
\begin{equation}
	\label{eq:forward_rgb}
	\resizebox{0.9\linewidth}{!}{
	\mbox{\fontsize{10}{12}\selectfont $
	\mathcal{L}_\text{rgb, d}^\text{for} = \sum\limits_{\mathbf{r}_\text{dst} \in \mathcal{R}_\text{C}} \left\| \left(1 - \hat{M}(\mathbf{r}_\text{dst})\right) \left( {\hat{\mathbf{C}}_c(\mathbf{r}_\text{dst}) - \bar{\mathbf{C}}(\mathbf{r}_\text{dst})} \right)\right\|_1,
	$ } } %
\end{equation}
where 
$\hat{\mathbf{C}}_c$ denotes a rendered color by $\mathbf{RF}_{\Theta_{c}}$ within $\mathcal{R}_\text{C}$.
We bilinearly interpolate input color based on the projected pixel's coordinates to estimate the color $\bar {\mathbf{C}}$.
$\hat {M}(\mathbf{r})$ is the estimated motion mask of $\mathbf{r}$ using the global mask module, detailed in Section~\ref{subsec:mask}.
$\mathbf{RF}_{\Theta_c}$ additionally uses the static regions of frames beyond the temporal window to maintain locality and incorporate richer view information.

\begin{figure}[tp]
	\centering
	\vspace{-2mm}
	\includegraphics[width=1\linewidth]{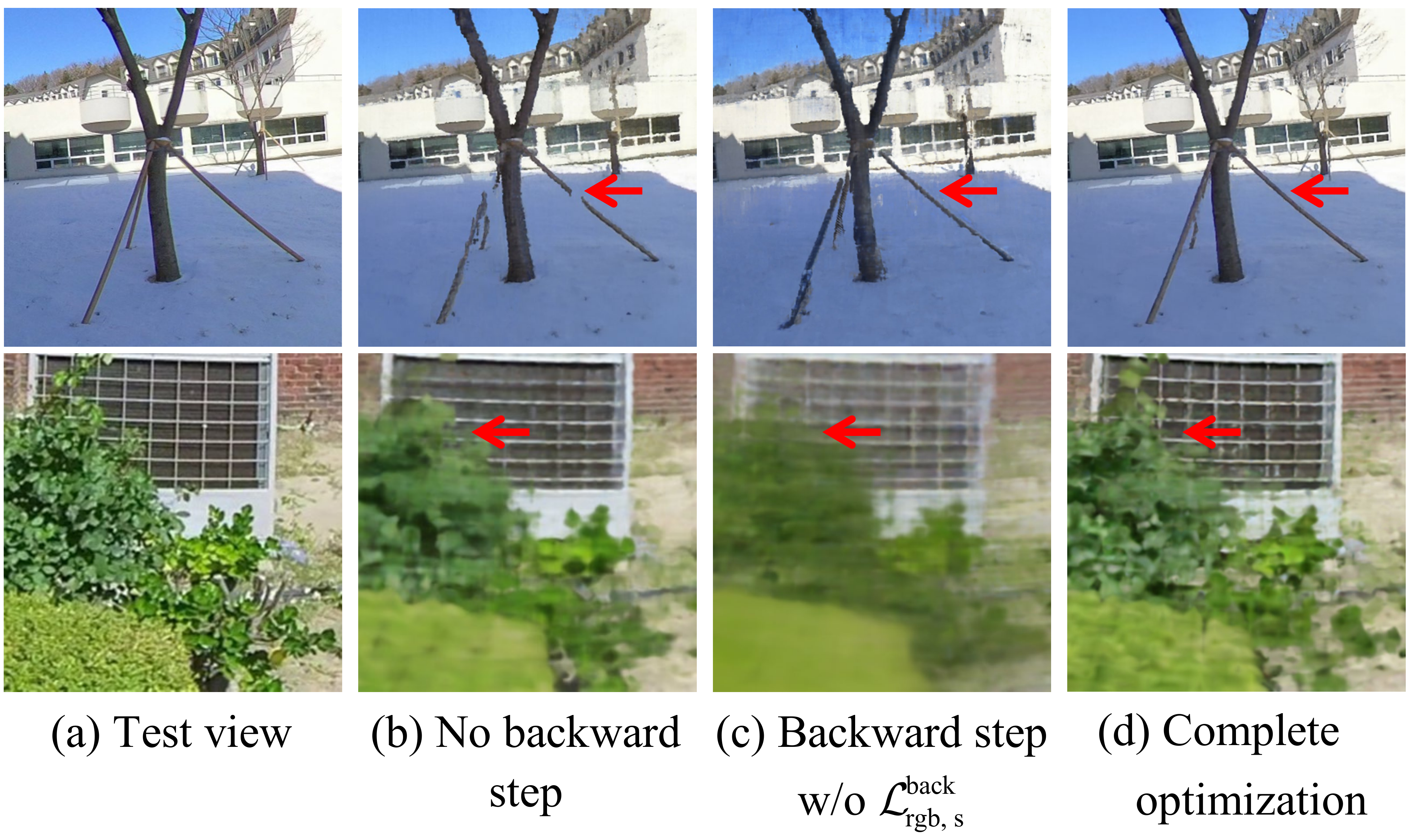}	
	\vspace{-8mm}
	\caption{\label{fig:br}
	Ablation study on the impact of the backward step. (b) Solely employing the forward step results in a blurred image. (c) Omitting the utilization of Eq.~\eqref{eq:backward_rgb_src} leads to overfitting on distant frames. (d) Our bidirectional optimization shows great quality in representing details.}
	\vspace{-5mm}
\end{figure}

\mparagraph{Backward step}
We illustrate the backward process in Figure~\ref{fig:brdf}(b). 
We cast a ray from $\mathbf{r}'_{\text{src}}$ on $w(p,j)$-th frames, which is used as the destination frame in the forward step, and render a depth value 
with a color value
through $\mathbf{RF}_{\Theta_{p}}$. 
We then supervise the rendered color by the color of the input frame as:
\begin{equation}
	\label{eq:backward_rgb_src}
	\resizebox{0.9\linewidth}{!}{
	\mbox{\fontsize{10}{12}\selectfont $
	\mathcal{L}_\text{rgb, s}^\text{back} = \sum\limits_{{\mathbf{r}'}_\text{src} \in \mathcal{R}} \left\| \left(1 - \hat{M}({\mathbf{r}'}_\text{src})\right) \left( {\hat{\mathbf{C}}_p({\mathbf{r}'}_\text{src}) - {\mathbf{C}}({\mathbf{r}'}_\text{src})} \right)\right\|_1,
	$ } } %
\end{equation}
where $\hat{\mathbf{C}}_p$ indicates a rendered color by $\mathbf{RF}_{\Theta_{p}}$.
Unless we use Eq.~\eqref{eq:backward_rgb_src}, $\mathbf{RF}_{\Theta_{p}}$ overfits to render distant frames while distorting adjacent views as shown in Figure~\ref{fig:br}(c). 

We get the projected pixel $\mathbf {p}'_\text{dst}$ on $w(c,i)$-th frame as follows:	
\begin{equation} \label{eq:p'_dst}
	\mathbf {p}'_\text{dst} = {\Pi} {\left( {{{[R\mid t]}_{w(p,j) \to w(c,i)}}{\Pi} {^{ - 1}\left(\mathbf{p}'_\text{src},\hat D_p({\mathbf{r}'_\text{src}}) \right)} } \right)}, 
\end{equation}
where the source and destination frame are reversed in Eq.~\eqref{eq:p_dst}.
We render a color with a depth through $\mathbf{RF}_{\Theta_{p}}$ and also render a depth value using $\mathbf{RF}_{\Theta_{c}}$ at $\mathbf{r}'_{\text{dst}}$.
The photometric loss for backward refinement supervises the color rendered by $\mathbf{RF}_{\Theta_{p}}$:
\begin{equation}
	\label{eq:backward_rgb}
	\resizebox{0.9\linewidth}{!}{
	\mbox{\fontsize{10}{12}\selectfont $
\mathcal{L}_\text{rgb, d}^\text{back} = \sum\limits_{\mathbf{r}'_\text{dst} \in \mathcal{R}_\text{P}} \left\| \left(1 - \hat{M}(\mathbf{r}'_\text{dst})\right) \left( {\hat{\mathbf{C}}_p(\mathbf{r}'_\text{dst}) - \bar{\mathbf{C}}(\mathbf{r}'_\text{dst})} \right)\right\|_1.
	$ } } %
\end{equation}
$\mathcal{R}_\text{P}$ denotes the valid ray bundles where the terms $\hat{D}_{c}(\mathbf{r})$ and $\hat{D}_{p}(\mathbf{r})$ are reversed in Eq.~\eqref{eq:forward_condition}, and $\bar{\mathbf{C}}(\mathbf{r}'_\text{dst})$ is a bilinear interpolation of ${\mathbf{C}}(\mathbf{r}'_{\text{dst}})$ based on  $\mathbf{p}'_{\text{dst}}$, a pixel origin of $\mathbf{r}'_{\text{dst}}$.

\begin{figure*}
	\centering
	\vspace{-7mm}	
	\includegraphics[width=0.79\textwidth]{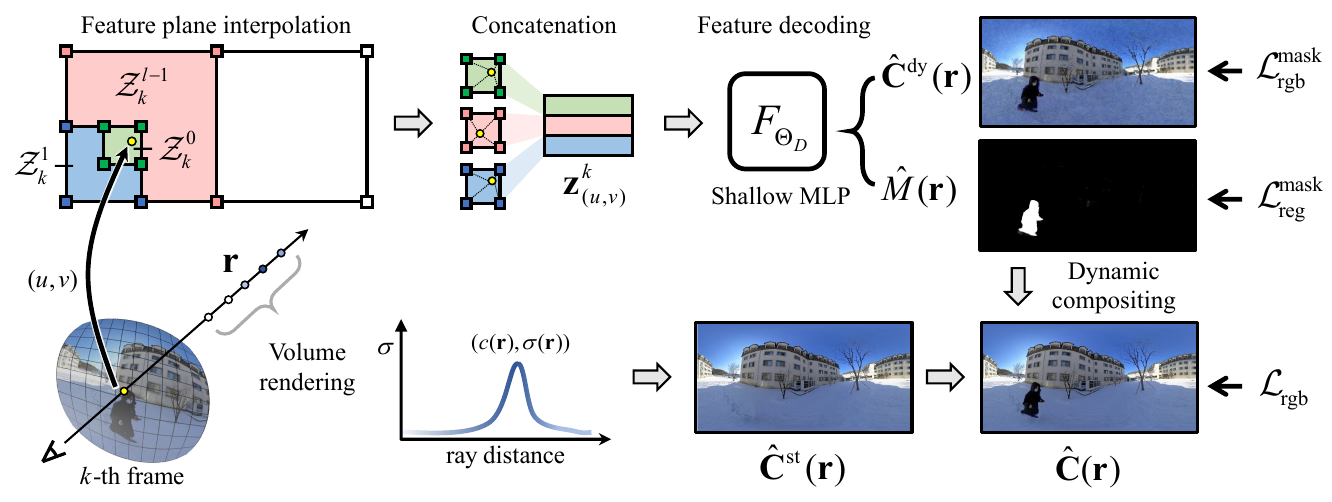} 
	\vspace{-5mm}
	\caption{\label{fig:overview}
	For motion mask prediction, we cast a ray $\mathbf{r}$ from the $k$-th frame and render the static structure $\hat{\mathbf{C}}^\text{st}(\mathbf{r})$ through volume rendering using radiance fields $\mathbf{RF}_{\Theta}$. We extract multi-resolution features of normalized $(u,v)$ by traversing feature plane set $\mathcal{Z}_k$ and concatenate them into a single code $\mathbf{z}^{k}_{(u,v)}$. We estimate dynamic color $\hat{\mathbf{C}}^\text{dy}(\mathbf{r})$ and motion mask $\hat {M}(\mathbf r)$, and render the final results $\hat{\mathbf{C}}(\mathbf{r})$ through dynamic compositing (Eq.~\eqref{eq:dynamic_compositing}).
	We jointly update the mask module $(\mathcal{Z}_t, {F_{{\Theta _{D}}}})$ with radiance fields $\mathbf{RF}_{\Theta}$ using L1 photometric loss. We supervise $\hat{\mathbf{C}}^\text{dy}(\mathbf{r})$ by $\tilde{\mathbf{C}}(\mathbf{r})$ for unique factorization (Eq.~\eqref{eq:mask_rgb}) and regularize the alpha of the mask (Eq.~\eqref{eq:mask_reg}).}%
	\vspace{-4mm}%
\end{figure*}

During the forward and backward steps, we use rays from frames close to the target RF block (source frame) and from frames far away from the block (destination frame). 
In this manner, we utilize four photometric loss functions, Eqs.~\eqref{eq:main_rgb}, \eqref{eq:forward_rgb}, \eqref{eq:backward_rgb_src}, and \eqref{eq:backward_rgb}.
Then, Eq.~\eqref{eq:main_rgb} is used to train $\mathbf{RF}_{\Theta_{c}}$ through both dynamic and static regions of the close frame. 
In this process, geometrically inconsistent areas are mapped to the mask module as dynamic objects.
Eq.~\eqref{eq:forward_rgb} refines $\mathbf{RF}_{\Theta_{c}}$ by far frame that have been used to train $\mathbf{RF}_{\Theta_{p}}$.  
As we already obtained the motion mask, we use only information from static regions to minimize the intervention of dynamic objects while refining.
We also apply this approach in the backward step, exclusively updating static regions during additional refinement of $\mathbf{RF}_{\Theta_{p}}$ (Eqs.~\eqref{eq:backward_rgb_src},~\eqref{eq:backward_rgb}).
Through this way, the overlapped regions in $\mathbf{RF}_{\Theta_{c}}$ and $\mathbf{RF}_{\Theta_{p}}$ are refined concurrently using distant frames.

\subsection{Motion Mask Prediction}\label{subsec:mask}
We use a set of neural feature planes for each frame to estimate a motion mask. 
To do so, we decode a feature code from a multi-resolution feature plane.
Our mask module is compatible with a pixel-wise ray marching NeRF setup and can be optimized jointly while rendering static objects.
Figure~\ref{fig:overview} provides an overview of our motion mask prediction.

\mparagraph{Mask module architecture}
We leverage a Bayesian learning framework for segmenting dynamic components inspired by DynIBaR~\cite{li2023dynibar} that combines IBRNet~\cite{wang2021ibrnet} with 2D CNN.
We create a low-resolution equirectangular feature plane set $\mathcal{Z}_k$ for each frame to handle delicate frame-dependent components.
The feature plane set comprises multi-resolution planes, denoted by $\mathcal{Z}_k=\left\{ \mathcal{Z}_k^1,\mathcal{Z}_k^2,\cdots,\mathcal{Z}_k^{L}\right\}$, where the height of each plane is $h_k^l = h_k^0/{2^{l - 1}}$, with $l \in \{ 1,\cdots,L\}$. 
This approach follows previous works~\cite{takikawa2021neural, muller2022instant, fridovich2023k, Turki_2022_CVPR} that have shown better spatial context-aware inference under a multi-resolution manner. We use 4 feature channels, $h_k^0=128$, and $L=4$ for all our experiments.

To render an omnidirectional image, a camera ray $\mathbf{r}$ is generated by multiplying the camera matrix with an equirectangular backprojected ray from pixel $\mathbf{p}(u,v)$ on the $k$-th frame. Before casting a ray from $\mathbf{p}$, we interpolate multi-resolution features at normalized $(u,v)$ by traversing $\mathcal{Z}_k$ and concatenate $L$ levels of features into a single code.
 We use a global, shallow, multi-layer perceptron (MLP) $F_{\Theta _{D}}$ to decode the feature code and obtain the color of dynamic objects $\hat {\mathbf{C}}^\text{dy}(\mathbf{r})$ with the alpha value of the motion mask $\hat M(\mathbf{r})$ at the ray $\mathbf{r}$.

\mparagraph{Motion mask optimization}
We compute the final color $\hat{\mathbf{C}}_m(\mathbf{r})$ by compositing dynamic results $\hat{\mathbf{C}}^\text{dy}(\mathbf{r})$ with static results $\hat{\mathbf{C}}^\text{st}_m(\mathbf{r})$, rendered by $\mathbf{RF}_{\Theta _{m}}$:
\begin{equation} \label{eq:dynamic_compositing}
	\hat{\mathbf{C}}_m(\mathbf{r}) = \hat{M}(\mathbf{r}) \hat{\mathbf{C}}^\text{dy}(\mathbf{r}) + (1-\hat{M}(\mathbf{r}))\hat{\mathbf{C}}^\text{st}_m(\mathbf{r}). 
\end{equation}
We update the mask module, consisting of the feature plane set and MLP, with radiance fields by propagating the L1 photometric loss of rendered colors (Section~\ref{subsec:brdf}).

The mask blended dynamic color in Eq.~\eqref{eq:dynamic_compositing} has a variety of combinations of a mask and a dynamic color.
When transient components have an intermediate alpha value due to the ambiguity in factorization, radiance fields try to compensate residuals and leave floating artifacts. 
Therefore, we supervise 
a dynamic color by Gaussian noise added input color $\tilde {\mathbf{C}}(\mathbf{r})$:
\begin{equation}\label{eq:mask_rgb}
	\mathcal{L}^\textrm{mask}_\text{rgb}=\sum\limits_{\mathbf{r} \in \mathcal{R}}  \left\| {\hat{\mathbf{C}}^\text{dy}(\mathbf r) - \tilde {\mathbf{C}}(\mathbf r)} \right\|_1, 
\end{equation}
which ensures a unique factorization while preventing our model from relying on the mask module to express fine details.
We regularize the motion mask by the total variation (TV) loss
and the binary loss for forcing the mask converged into a binary value while smoothing it as: 
\begin{equation}\label{eq:mask_reg}
	\mathcal{L}^\textrm{mask}_\text{reg}=\mathcal{L}^\textrm{mask}_\text{TV}+\mathcal{L}^\textrm{mask}_\text{bin}. 
\end{equation}
Refer to the supplementary material for details on the mask regularizers.

\subsection{Progressive Optimization}
\label{sec:progressive_optimization}
We optimize $\bold{RF}_{\Theta}$ blocks, camera poses, feature plane sets, and mask MLP by sliding a window over the input video. We insert frames into the window and optimize poses using $\bold{RF}_{\Theta_c}$ (Eq.~\eqref{eq:main_rgb}). 
If the poses fall outside the contraction range of $\bold{RF}_{\Theta_c}$, currently targeted radiance field, we move to the refining step. 
In the refining step, we perform bidirectional optimization by simultaneously updating the $\bold{RF}_{\Theta_c}$ and the randomly selected $\bold{RF}_{\Theta_p}$ that has been previously optimized. 
We use LocalRF's optical flow loss and normalized monocular depth supervision for robust pose estimation. 
To render novel views, we search for the nearest frame from the given viewpoints. If a frame is used to train two adjacent radiance blocks, we blend the results based on their position among overlapped frames.

\section{Experimental Results}
\label{sec:experiments}
Our method takes 12 hours for 125 frames on a machine equipped with a single NVIDIA A6000 GPU and an Intel Xeon Silver 4214R 2.40 GHz CPU with 256 GB RAM. 
Refer to the supplemental material for further implementation details. 
To evaluate OmniLocalRF in large-scale view synthesis from 360{\degree} videos, we compare our approach against Mip-NeRF360~\cite{barron2022mipnerf360} and EgoNeRF~\cite{Choi_2023_CVPR}, known for strong performance in omnidirectional view synthesis.
Additionally, we also compare our approach with D$^2$NeRF~\cite{wu2022d2nerf} and RobustNeRF~\cite{sabour2023robustnerf}, both of which aim to reconstruct static structures from dynamic videos.
Since these methods require precomputed camera pose, we utilize the pose estimated by OpenVSLAM~\cite{sumikura2019openvslam} as camera priors. 
We also compare ours with LocalRF~\cite{meuleman2023localrf} that self-calibrate poses during view synthesis. 
For a comprehensive comparison, we evaluate both LocalRF and our method under two conditions: one with camera priors provided for pose refinement and the other without camera priors, starting from scratch and estimating the poses during view synthesis.

\subsection{Dataset and Metrics}
\label{sec:dataset}
We capture and provide a new dataset from six outdoor scenes captured with an Insta360 camera, consisting of 5760$\times$2880 resolution 30\,fps 360$\degree$ videos each. 
These are casual videos designed to capture backgrounds that include a photographer and dynamic objects like pedestrians.
Considering reasonable training time and memory size, we use half-resolution input images. However, in the case of D$^2$NeRF, we have to use a quarter of the spatial resolution for rendering due to GPU memory constraints of 48\,GB.
We use a total of 125 images, taking every fourth frame from the first 500 frames. Test frames are selected for every tenth frame among 125 images, including the first image, resulting in 112 training images with 13 test images.
For all test images in every scene, we manually create motion masks for validation. 
We then compare the rendered results with the dynamic objects masked ground truth and report PSNR, SSIM~\cite{wang2004ssim}, and LPIPS~\cite{zhang2018lpips}. 
We also measure weighted-to-spherically uniform PSNR and SSIM~\cite{sun2017weighted} (PSNR$^\text{WS}$ and SSIM$^\text{WS}$), taking into account the distortion near the pole in equirectangular images.

\begin{figure}
	\centering
	\includegraphics[width=1\linewidth]{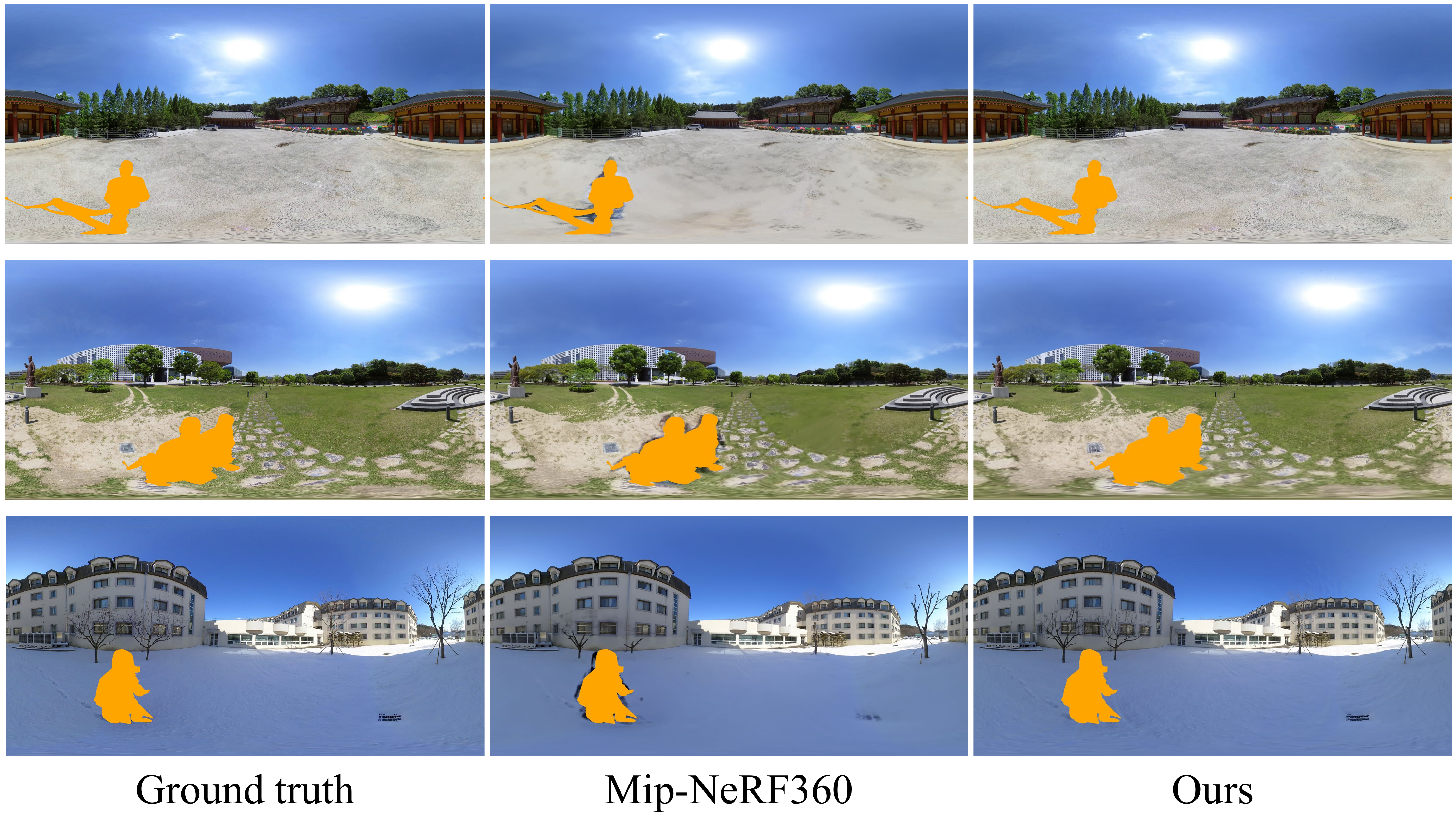}%
	\vspace{-4mm}%
	\caption{\label{fig:masked_results}
	Masked results for evaluation in the real dataset. We manually create motion masks for the test views and compute the metrics after inpainting masked regions with gray. 
	While our method masks the dynamic areas robustly, conventional neural rendering methods still exhibit residual artifacts due to temporal and spatial inconsistency.}
	\vspace{-4mm}
\end{figure}

We generate three synthetic 360{\degree} videos, each containing dynamic objects floating within the scene for evaluating pose estimation accuracy by absolute trajectory error (ATE) and relative pose error (RPE) between ground truth poses as standard visual odometry metrics~\cite{kopf2021robust, sturm2012benchmark, zhang2018tutorial}.
The synthetic videos have a resolution of 2880$\times$1440, consisting of 125 images each. Out of these, 13 images are allocated for testing, following the procedure used with the real dataset. 
Additionally, we provide view synthesis results of existing methods and our method for comprehensive comparisons.

\begin{figure*}
	\centering
	\includegraphics[width=1.0\textwidth]{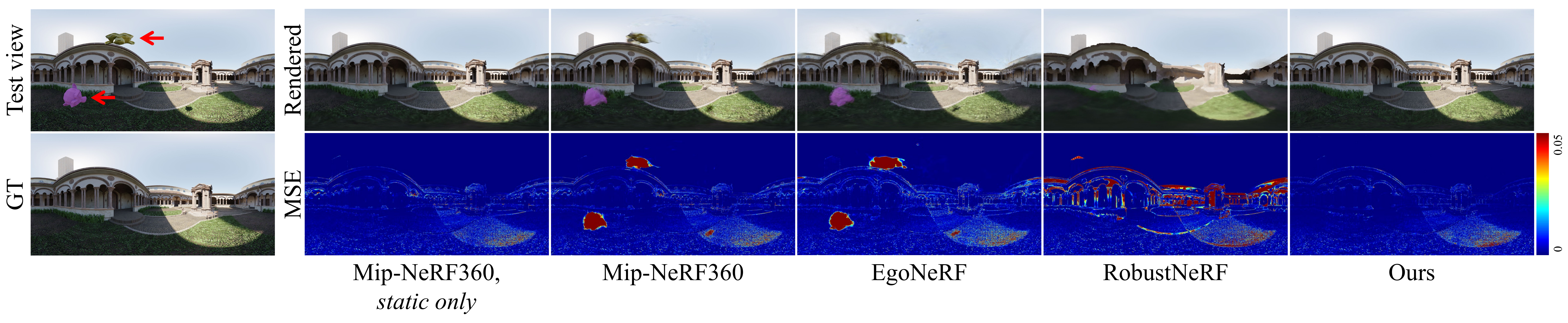}\\	
	\vspace{-5mm}
	\caption{\label{fig:comparison_syn}
	An example of qualitative comparisons on the \textit{Lone Monk} scene within the synthetic dataset. 
	We additionally address results of Mip-NeRF360 from static-only videos (\textit{static only}) that do not include dynamic objects at the first column. Our results demonstrate the effective removal of dynamic objects while preserving details in the expansive scene.}
	\vspace{-4mm}
\end{figure*} 
\begin{figure*}
	\centering
	\includegraphics[width=1.0\textwidth]{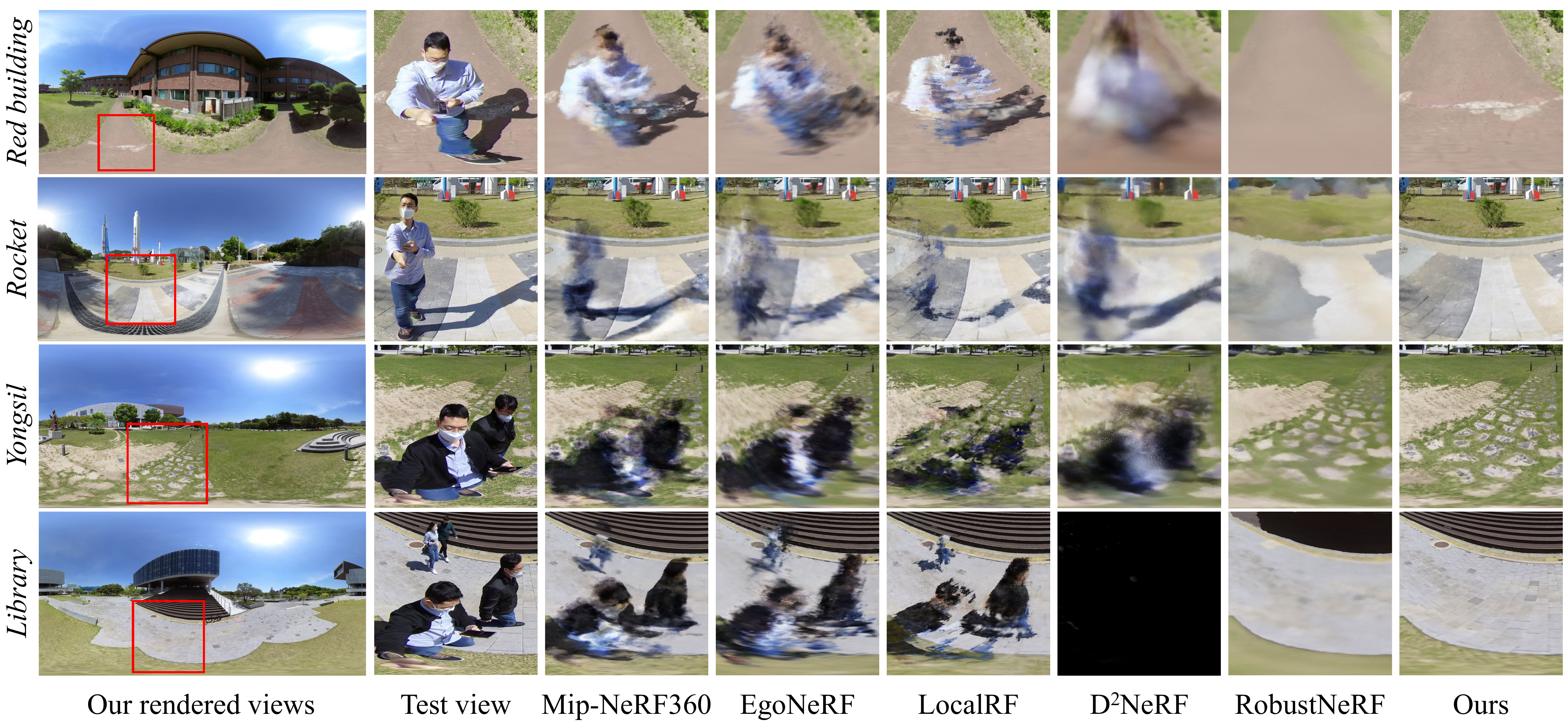}\\
	\vspace{-4mm}%
	\caption{\label{fig:comparison_qual_v3}
	Qualitative comparisons on the real dataset. Our method can render locally unobserved areas with fine details throughout the bidirectional optimization with a self-supervised mask module. D$^2$NeRF sometimes encounters challenges in distinguishing between static backgrounds and dynamic objects, which results in the omission of certain regions in the \textit{Library} scene. Refer to the supplemental video.}
	\vspace{-4mm}
\end{figure*} 

\subsection{Quantitative Evaluation}
Dynamic objects in test images are excluded using manually created masks when computing metrics. As both LocalRF and our method can estimate camera poses during view synthesis, we present the metrics for models trained with and without preprocessed camera poses by OpenVSLAM~\cite{sumikura2019openvslam}.
According to Tables~\ref{tb:quantative_syn} and~\ref{tb:quantative}, scalable models like Mip-NeRF360, EgoNeRF, and LocalRF showed relatively high performance.
D$^2$NeRF and RobustNeRF, which target reconstructing static structures from dynamic videos, achieve lower scores due to a limited model capacity.
To ensure fair comparisons, we exclude the transient data of test views. However, dynamic objects frequently appear outside the masks as artifacts, leading to a decrease in the metrics, as illustrated in Figure~\ref{fig:masked_results}. 
As a result, our method outperforms existing methods thanks to its effective dynamic object removal and locality even without given camera poses. 

\begin{table}[!h]
	\tiny
	\centering
	\caption{\label{tb:quantative_syn} 
		Quantitative comparisons of the synthetic dataset. The averages of the metrics are measured across three synthetic scenes. %
		Refer to Figure~\ref{fig:comparison_syn} for qualitative comparisons.
	}
	\vspace{-3mm}
	\resizebox{\linewidth}{!}{
	\begin{tabular}
		{l|ccccc}
		\thickhline
		&PSNR $\uparrow$	&PSNR$^\text{WS}$ $\uparrow$	&SSIM $\uparrow$		&SSIM$^\text{WS}$ $\uparrow$	&LPIPS $\downarrow$\\
		\hline
		Mip-NeRF360~\cite{barron2022mipnerf360},	\textit{static only}	&29.48	&29.85				&0.8625				&0.8628				&0.2699				 \\
		Ours	wo/ pose, \textit{static only}	&30.29	&30.24				&0.8679			&0.8681			&0.2561				 \\
		\hline
		Mip-NeRF360~\cite{barron2022mipnerf360}		&\cellcolor[HTML]{FFFFC7}25.59	&\cellcolor[HTML]{FFFFC7}25.32				&\cellcolor[HTML]{FFFFC7}0.8455	&\cellcolor[HTML]{FFFFC7}0.8464				&0.2973				 \\
		EgoNeRF~\cite{Choi_2023_CVPR}				&23.99	&23.67				&0.8044				&0.7951 			&0.3949				 \\
		LocalRF~\cite{meuleman2023localrf} w/ pose	&25.50	&25.31				&0.8454				&0.8427		&\cellcolor[HTML]{FFFFC7}0.2897			 \\
		D$^2$NeRF~\cite{wu2022d2nerf}				&19.91	&19.43 				&0.6212				&0.5929			 	&0.6298				 \\
		RobustNeRF~\cite{sabour2023robustnerf}		&20.59 	&19.79 				&0.7326 			&0.7096				&0.4734				 \\		
		Ours w/ pose								&\cellcolor[HTML]{ffdcb2}29.76 	&\cellcolor[HTML]{ffdcb2}29.68 				&\cellcolor[HTML]{ffdcb2}0.8633			&\cellcolor[HTML]{ffdcb2}0.8628		&\cellcolor[HTML]{ffdcb2}0.2624	 \\
		\hline
		LocalRF~\cite{meuleman2023localrf} wo/ pose	&25.22	&25.02	 			&0.8389				&0.8354				&0.2949				 \\
		Ours wo/ pose								&\cellcolor[HTML]{FFCCC9}29.93	&\cellcolor[HTML]{FFCCC9}29.85 				&\cellcolor[HTML]{FFCCC9}0.8648 			&\cellcolor[HTML]{FFCCC9}0.8648				&\cellcolor[HTML]{FFCCC9}0.2610				 \\
		\thickhline
	\end{tabular}
	}
	\vspace{-3mm}
\end{table}

\begin{table}
	\scriptsize
	\centering
	\caption{\label{tb:quantative} 
		Quantitative comparisons of the real dataset. We report the averages of the metrics measured across six real scenes. Refer to Figure \ref{fig:comparison_qual_v3} for qualitative comparisons.}	
	\vspace{-3mm}
	\resizebox{\linewidth}{!}{
	\begin{tabular}
		{l|ccccc}
		\thickhline
		&PSNR $\uparrow$	&PSNR$^\text{WS}$ $\uparrow$	&SSIM $\uparrow$		&SSIM$^\text{WS}$ $\uparrow$	&LPIPS $\downarrow$\\
		\hline
		Mip-NeRF360~\cite{barron2022mipnerf360}		&\cellcolor[HTML]{FFFFC7}26.88	&\cellcolor[HTML]{FFFFC7}26.44				&\cellcolor[HTML]{FFFFC7}0.8094				&0.7977				&0.3585				 \\
		EgoNeRF~\cite{Choi_2023_CVPR}				&25.95	&25.38				&0.7609				&0.7424 			&0.4383				 \\
		LocalRF~\cite{meuleman2023localrf} w/ pose	&26.56 	&26.22				&0.8041				&0.7966		&0.3471				 \\
		D$^2$NeRF~\cite{wu2022d2nerf}				&20.95	&20.34 				&0.6105				&0.5829			 	&0.5100				 \\
		RobustNeRF~\cite{sabour2023robustnerf}		&20.78 	&19.56 				&0.7093 			&0.6679				&0.4864				 \\		
		Ours w/ pose								&\cellcolor[HTML]{ffdcb2}27.72 &\cellcolor[HTML]{ffdcb2}27.09 				&\cellcolor[HTML]{FFCCC9}0.8171 				&\cellcolor[HTML]{ffdcb2}0.8085			&\cellcolor[HTML]{ffdcb2}0.3299 \\
		\hline
		LocalRF~\cite{meuleman2023localrf} wo/ pose	&26.56	&26.23	 			&0.8034				&\cellcolor[HTML]{FFFFC7}0.7984				&\cellcolor[HTML]{FFFFC7}0.3410				 \\
		Ours wo/ pose								&\cellcolor[HTML]{FFCCC9}27.73 &\cellcolor[HTML]{FFCCC9}27.13			&\cellcolor[HTML]{ffdcb2}0.8165		&\cellcolor[HTML]{FFCCC9}0.8088			&\cellcolor[HTML]{FFCCC9}0.3297				 \\		\thickhline
	\end{tabular}
	}
		\vspace{-6mm}
\end{table}

\subsection{Qualitative Evaluation}
In Figures~\ref{fig:comparison_syn} and~\ref{fig:comparison_qual_v3}, we conduct a qualitative comparison of our OmniLocalRF with the baseline methods.
Our method effectively mitigates ghosting artifacts while preserving locality, allowing us to represent details in large-scale real-world data, even in locally occluded regions.
Methods like EgoNeRF, Mip-NeRF360, and LocalRF, which lack dynamic handling, face challenges when reconstructing static scenes, relying solely on geometric consistency across input data.
Even though their masked metrics indicate high performance, these methods suffer from ghosting artifacts, making them less suitable for practical applications that demand high-quality static view synthesis.

Since D$^2$NeRF parameterizes radiance along spatial and temporal domains with a single NeRF module, it often fails to separate dynamic components. It renders blurry results because of the large spatial and time complexity of input videos.
RobustNeRF employs the iteratively reweighted least squares (IRLS) approach for patch-wise outlier down-weighting, which decouples transient artifacts from geometrically consistent structures through adaptive loss reweighting. 
However, in the omnidirectional videos, the overall geometry reconstruction quality is compromised due to the large scene scale, making it challenging to distinguish static from dynamic objects based on the photometric error over patches. 
For a fair comparison, we additionally train RobustNeRF for twice the number of iterations compared to Mip-NeRF360, taking into account the slowdown caused by the IRLS approach, but still fail to represent fine details effectively.

\subsection{Pose Estimation Comparison }
\begin{figure}
	\centering
	\vspace{-3mm}%
	\includegraphics[width=1\linewidth]{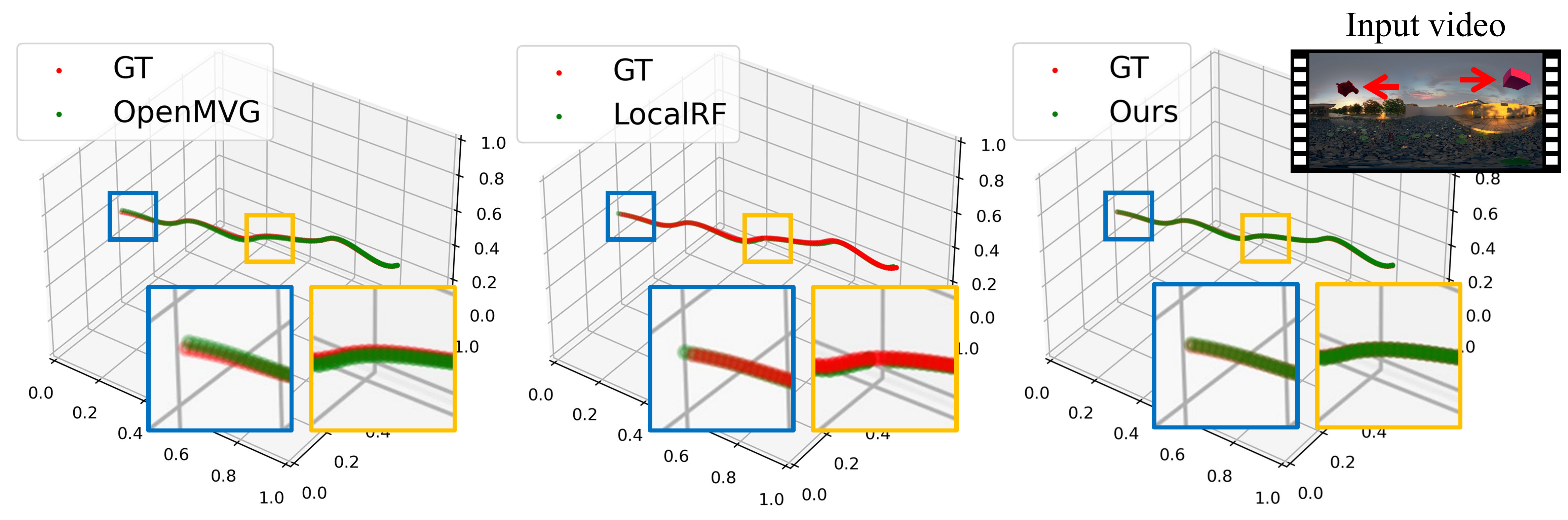}%
	\vspace{-4mm}%
	\caption{\label{fig:result_poses}
	An example of pose comparisons on the \textit{Pavillion} scene including dynamic objects. Our method shows robustness in estimating camera pose in the presence of dynamic objects.}
	\vspace{-3mm}
\end{figure}

We compare our method with OpenMVG~\cite{moulon2017openmvg}, an SfM-based pose estimator, and LocalRF in Figure~\ref{fig:result_poses}.
We report the relative rotation error (RPE$_r$), relative translation error (RPE$_t$), and ATE.
We also estimate the camera trajectories from the ground truth videos to eliminate the influence of dynamic objects.
We align the estimated pose with the ground truth, addressing scale factor and rotation discrepancies, before computing ATE.
As shown in Table~\ref{tb:pose}, the pose optimized through our approach is robust in the presence of dynamic objects, showing similar results in the static-only videos.

\begin{table}
	\scriptsize
	\centering
	\caption{\label{tb:pose} 
		Comparisons of pose accuracy on the synthetic dataset. We average the results from three scenes.}
	\vspace{-2mm}
	\resizebox{\linewidth}{!}{
	\begin{tabular}
		{l|ccc}
		\thickhline		
		& RPE$_r$ $\downarrow$ 	& RPE$_t$ $\downarrow$	&ATE $\downarrow$\\
		\hline
		OpenMVG~\cite{moulon2017openmvg}	&\cellcolor[HTML]{FFFFC7}0.10761	&\cellcolor[HTML]{FFFFC7}0.01799	&\cellcolor[HTML]{ffdcb2}0.00218 \\
		LocalRF~\cite{meuleman2023localrf} &\cellcolor[HTML]{ffdcb2}0.10404	&\cellcolor[HTML]{ffdcb2}0.00096	&\cellcolor[HTML]{FFFFC7}0.00376	\\
		Ours	&\cellcolor[HTML]{FFCCC9}0.10398	&\cellcolor[HTML]{FFCCC9}0.00081	&\cellcolor[HTML]{FFCCC9}0.00165	\\
		\hline
		OpenMVG~\cite{moulon2017openmvg}, \textit{static only} &0.10706 	&0.01796	&0.00187 \\	
		LocalRF~\cite{meuleman2023localrf}, \textit{static only}	&0.10398	&0.00071	&0.00208	\\		
		Ours, \textit{static only}		&0.10399	&0.00074	&0.00165	\\
		\thickhline
	\end{tabular}
	}\vspace{-4mm}
\end{table}

\subsection{Ablation Study}\label{ablation}
We conduct an ablation study for our proposed mask module and bidirectional optimization:
(a) Removing Eq.~\eqref{eq:forward_rgb} degrades overall quality as the local blocks use fewer input images for training. 
(b) The omission of source rays' photometric supervision during the backward step (Eq.~\eqref{eq:backward_rgb_src}) significantly degrades the model as it tends to converge on reconstructing distant regions as described in Figure~\ref{fig:br}(c).
(c) We observe that incorporating distant frames through the backward step enables the model to capture and refine detailed information.
(d) Geometrical inconsistencies in the temporal domain of dynamic objects often lead to the introduction of larger artifacts compared to the regions masked out during test views, as shown in Figure~\ref{fig:masked_results}. 
Consequently, the absence of the mask module adversely affects both the qualitative and quantitative aspects of the results.
(e) The exclusion of Eq.~\eqref{eq:main_rgb} during mask optimization results in intermediate alpha values within the motion mask due to the ambiguity in factorization and leaves floating artifacts, which yields lower metrics than the entire model.
Table~\ref{tb:ablation} quantitatively compares results.

\begin{table}	
	\small
	\centering
	\vspace{-3mm}
	\caption{\label{tb:ablation} 
		Ablation study results of our model. Reported metrics are averaged over the six scenes on our real dataset (Section~\ref{ablation}).
	}\vspace{-3mm}
	\resizebox{\linewidth}{!}{
	\begin{tabular}
		{l|ccc}
		\thickhline
		&PSNR $\uparrow$				&SSIM $\uparrow$				&LPIPS $\downarrow$\\
		\hline
		(a) No forward step		&\cellcolor[HTML]{ffdcb2}27.64			&\cellcolor[HTML]{ffdcb2}0.8127			 	&\cellcolor[HTML]{ffdcb2}0.3356			 \\
		(b) Backward step w/o $\mathcal{L}^{\text{back}}_{\text{rgb, s}}$	&26.23	&0.7800				&0.3866		\\		
		(c) No backward step &\cellcolor[HTML]{FFFFC7}27.63	&\cellcolor[HTML]{FFFFC7}0.8086				&\cellcolor[HTML]{FFFFC7}0.3447	 \\
		(d) No mask module	&26.08 	&0.7876 					&0.3630	 \\
		(e) No mask photometric supervision	&25.27 	&0.7894 					&0.3593	 \\
		\hline
		Complete model				&\cellcolor[HTML]{FFCCC9}27.73			&\cellcolor[HTML]{FFCCC9}0.8165			&\cellcolor[HTML]{FFCCC9}0.3297					 \\
		\thickhline
	\end{tabular}	
}\vspace{-6mm}
\end{table}

\section{Discussion and Conclusions}
\label{sec:conclusion}
We have presented OmniLocalRF, a novel method for omnidirectional view synthesis in dynamic 360{\degree} videos. Our approach integrates LocalRF with a mask module and bidirectionally refines distant NeRF blocks to remove dynamic artifacts and fill in occluded regions, resulting in accurate static structure reconstruction while preserving fine details within large scenes. Our method also accurately estimates camera trajectories during view synthesis, making it suitable for various applications such as street viewers and augmented reality environments.

Our model can synthesize static structures from 360{\degree} videos without motion masks and camera priors. However, it faces the usual challenges associated with neural rendering-based view synthesis. For instance, it is unable to inpaint regions that are completely occluded in the videos because NeRF-based models are trained using photometric loss between input images. To overcome this limitation, incorporating perceptual loss~\cite{spinnerf} or generative models~\cite{poole2022dreamfusion, weder2023removing}, such as stable diffusion, can be helpful.

We use linear interpolation in equirectangular space, which has grids with the same size of zenith and azimuth angles, to predict motion masks. Operating in this space can mitigate data redundancy compared to utilizing an undistorted cube map.
However, this space can lead to inefficient oversampling near polar regions in mask predictions. 
To address this issue, we could use uniformly sampled spherical grids in future works.

Even though our model globally refines the local blocks based on photometric error, we do not deal with the global bundle adjustment and loop closure for pose estimation, which are used in completed SLAM systems. Adding these components to our approach would enable more robust and accurate pose estimation.
While our model is capable of generating static structures from 360{\degree} videos, it faces several challenges that require further refinement.

\appendix
\section*{Acknowledgements}
\noindent Min H.~Kim acknowledges the MSIT/IITP of Korea (RS-2022-00155620, 2022-0-00058, and 2017-0-00072), LIG, and Samsung Electronics.

\clearpage
{
    \small
    \bibliographystyle{ieeenat_fullname}
    \bibliography{OmniLocalRF-reference}
}

\maketitlesupplementary

\section{Implementation Details}
\label{sec:details}
We schedule learning rate and upsampling based on frames per RF block to balance iteration over blocks, following LocalRF~\cite{meuleman2023localrf}.

\subsection{Regularizers for Motion Mask}
We employ a total variation loss $\mathcal{L}^\textrm{mask}_\text{TV}$ and a binary entropy~\cite{yuan2021star} like function $\mathcal{L}^\textrm{mask}_\text{bin}$ to regularize the motion mask.
Total variation is commonly utilized for smoothing a target and we adapt it as follows:
\begin{equation} \label{eq:reg_tv}
	\mathcal{L}^\textrm{mask}_\text{TV} = {\Delta}^2 M_{x} + {\Delta}^2 M_{y} + 0.1 {\cdot} {\Delta}^2 M_{t},
\end{equation}
where $M_x$, $M_y$ and $M_t$ denote the predicted mask value along the $x$, $y$ and temporal axes.
Here, ${\Delta}^2$ represents a squared difference between neighboring values along the axes. 
Considering object movement over time, we impose a relatively small weight along the temporal axis.
Additionally, we enforce the mask to converge into a binary value using 
\begin{equation} \label{eq:reg_tv}
	\mathcal{L}^\textrm{mask}_\text{bin} = {M(r)} ^ 2 {(1-M(r))}^2.
\end{equation}
The function exhibits a similar shape to the binary entropy regularizer, used as a slightly modified form in D$^2$NeRF~\cite{wu2022d2nerf} for the same purpose.

\subsection{Optimization Details}
In our method, we insert frames every $100$ iterations. 
The insertion process stops when the distance between the inserted frames and the center of the current radiance block exceeds $1$. 
This distance is equivalent to the radius of the contraction fields. 
After that, we optimize the blocks until the iteration reaches the number of frames in the current block multiplied by $600$. 
The resolution of RF blocks increases exponentially from $64^3$ to $640^3$ at iterations $[100, 150, 200, 250, 300]$ times the number of frames in the block. 
Once we optimize the block, a new radiance block is created to overlap frames with the previous block for further optimization. 

For pose estimation, we follow the camera parameterization of LocalRF. 
The rotation matrix comprises two perpendicular vectors with their cross-product, and the translation matrix is defined in Euclidean space. During training, the camera pose is updated by photometric error while training RF blocks. Test views are solely used to optimize the pose without contributing to the training of RF blocks.
We train our model for each scene using $155$K iterations. 
Our method utilizes adaptive sampling, as proposed in TensoRF~\cite{chen2022tensorf}. 
The number of samples exponentially increases from $219$ to $2,214$ at the upsampling iteration. 
We initiate a backward step after completing the upsampling of the current radiance block. 
We do not perform a backward step during frame insertion as it could contaminate the previous radiance fields due to the inaccurate pose of the current frames.

\begin{figure}[pt]
	\centering
	\includegraphics[width=1\linewidth]{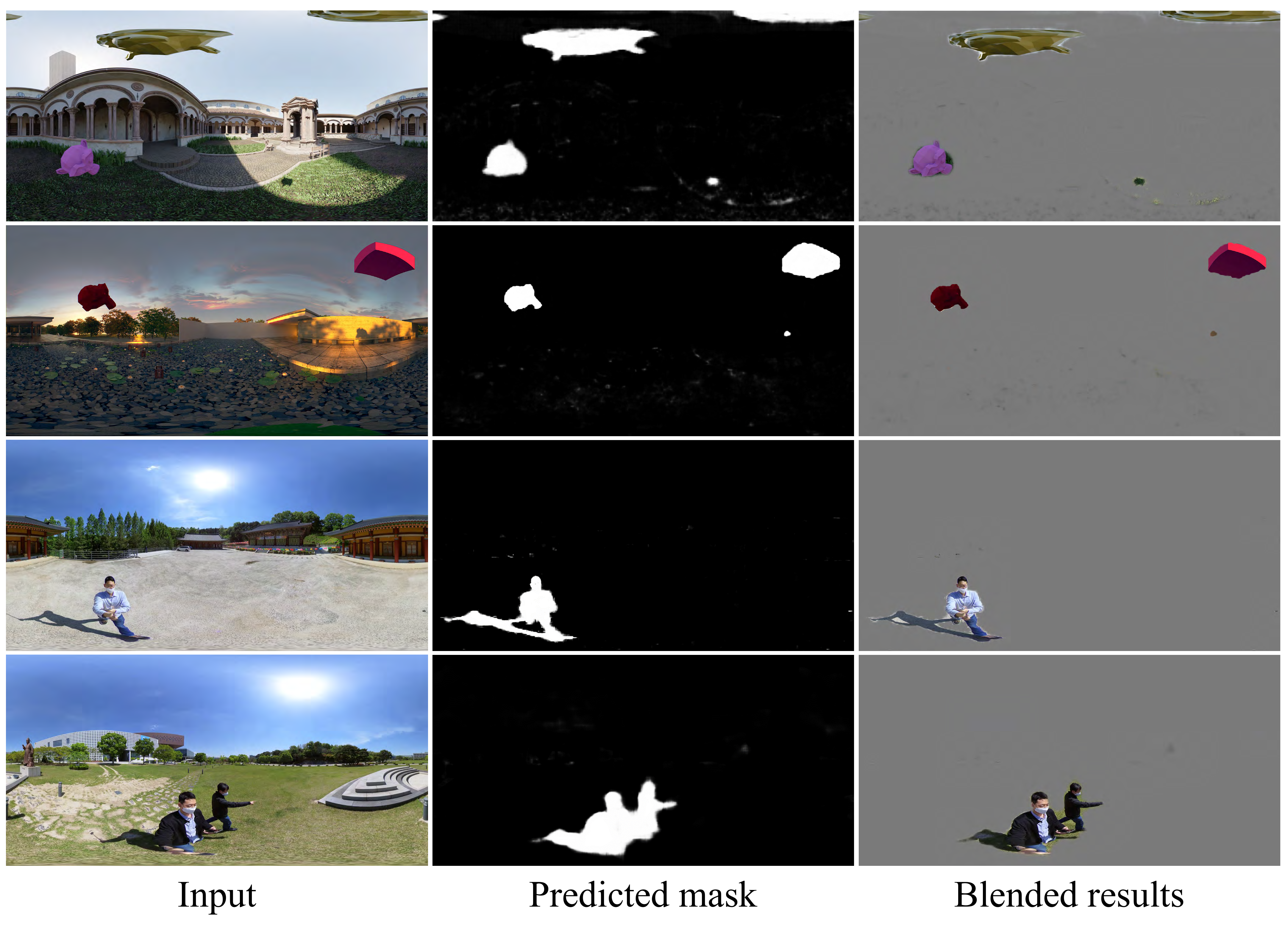}%
	\vspace{-2mm}
	\caption{\label{fig:mask_results}
		Predicted motion mask results. The second column is estimation results from our mask module, and the last column shows blended results with the input images.}
	\vspace{3mm}
	\centering
	\includegraphics[width=1\linewidth]{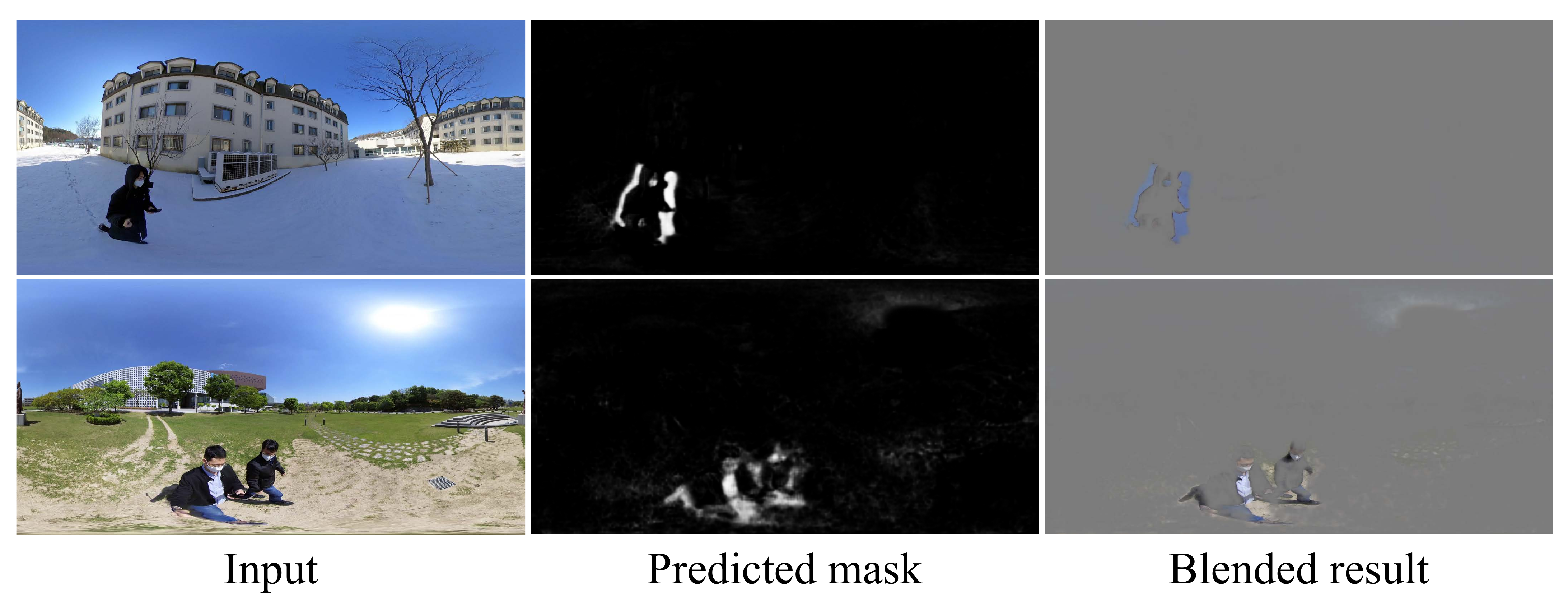}%
	\vspace{-2mm}%
	\caption{\label{fig:mask_wo_rgb}
		Predicted mask results without photometric supervision of the ground truth. Unresolving factorization ambiguity often leads to an intermediate alpha value of the motion mask, leaving dynamic artifacts on the static geometry.}
\end{figure}

\begin{table*}[th]
	\vspace{-5mm}
	\scriptsize
	\centering
	\caption{\label{tb:quantative_real} 
		Quantitative comparisons of the real dataset. Refer to Figures \ref{fig:comparison_qual_dormitory}, \ref{fig:comparison_qual_rocket} and \ref{fig:comparison_qual_temple} for qualitative comparisons.}	
	\vspace{-2mm}
	\resizebox{\linewidth}{!}{
		\begin{tabular}
			{l|ccccc|ccccc|ccccc}
			\thickhline
			&\multicolumn{5}{c|}{\textit{Dormitory}} &\multicolumn{5}{c|}{\textit{Library}} &\multicolumn{5}{c}{\textit{Red building}}  \\
			&PSNR 	&PSNR$^\text{WS}$	&SSIM				&SSIM$^\text{WS}$	&LPIPS
			&PSNR 	&PSNR$^\text{WS}$	&SSIM				&SSIM$^\text{WS}$	&LPIPS
			&PSNR 	&PSNR$^\text{WS}$	&SSIM				&SSIM$^\text{WS}$	&LPIPS\\
			\hline
			Mip-NeRF360~\cite{barron2022mipnerf360}		 &25.95 &24.70 &0.8564 &0.8220 &0.3837 &\cellcolor[HTML]{FFFFC7}28.02 &\cellcolor[HTML]{FFFFC7}27.75 &0.8485 &0.8489 &0.3280 &\cellcolor[HTML]{FFCCC9}26.60 &\cellcolor[HTML]{FFCCC9}25.77 &0.7998 &\cellcolor[HTML]{FFFFC7}0.7672 &0.3719	 \\
			EgoNeRF~\cite{Choi_2023_CVPR}				 &26.11 &24.80 &0.8404 &0.7988 &0.4190 &27.05 &26.49 &0.7977 &0.7884 &0.4141 &25.55 &24.64 &0.7557 &0.7078 &0.4568 	 \\
			LocalRF~\cite{meuleman2023localrf} w/ pose	 &\cellcolor[HTML]{FFFFC7}27.20 &\cellcolor[HTML]{FFFFC7}26.12 &\cellcolor[HTML]{FFFFC7}0.8826 &\cellcolor[HTML]{FFFFC7}0.8589 &\cellcolor[HTML]{FFFFC7}0.3180 &27.72 &27.47 &\cellcolor[HTML]{FFCCC9}0.8547 &\cellcolor[HTML]{FFCCC9}0.8540 &\cellcolor[HTML]{FFCCC9}0.2930 &25.68 &24.90 &0.7860 &0.7492 &0.3743 	 \\
			D$^2$NeRF~\cite{wu2022d2nerf}				 &24.96 &21.72 &0.7887 &0.731 &0.4687 &4.44 &4.90 &0.0733 &0.0779 &0.7084 &25.55 &25.16 &0.7487 &0.7056 &0.4441 	 \\
			RobustNeRF~\cite{sabour2023robustnerf}		 &19.44 &17.98 &0.794 &0.7330 &0.4850 &20.78 &19.23 &0.7338 &0.6908 &0.4738 &19.91 &18.49 &0.7098 &0.6409 &0.4974 	 \\
			Ours w/ pose								&\cellcolor[HTML]{ffdcb2}29.10 &\cellcolor[HTML]{ffdcb2}27.86 &\cellcolor[HTML]{ffdcb2}0.8956 &\cellcolor[HTML]{ffdcb2}0.8748 &\cellcolor[HTML]{ffdcb2}0.2992 &\cellcolor[HTML]{ffdcb2}29.61 &\cellcolor[HTML]{ffdcb2}28.81 &\cellcolor[HTML]{ffdcb2}0.8544 &\cellcolor[HTML]{FFFFC7}0.8535 &\cellcolor[HTML]{ffdcb2}0.2970 &\cellcolor[HTML]{FFFFC7}26.03 &25.18 &\cellcolor[HTML]{FFFFC7}0.8010 &0.7641 &\cellcolor[HTML]{FFFFC7}0.3534 	 \\
			\hline
			LocalRF~\cite{meuleman2023localrf} wo/ pose	 &26.59 &25.48 &0.8762 &0.8493 &0.3227 &27.70 &27.35 &0.8399 &0.8408 &0.3096 &\cellcolor[HTML]{ffdcb2}26.56 &\cellcolor[HTML]{ffdcb2}25.75 &\cellcolor[HTML]{ffdcb2}0.8040 &\cellcolor[HTML]{FFCCC9}0.7728 &\cellcolor[HTML]{FFCCC9}0.3452 \\
			Ours wo/ pose								&\cellcolor[HTML]{FFCCC9}29.23 &\cellcolor[HTML]{FFCCC9}27.99 &\cellcolor[HTML]{FFCCC9}0.8971 &\cellcolor[HTML]{FFCCC9}0.8767 &\cellcolor[HTML]{FFCCC9}0.2977 &\cellcolor[HTML]{FFCCC9}29.69 &\cellcolor[HTML]{FFCCC9}28.90 &\cellcolor[HTML]{FFFFC7}0.8540 &\cellcolor[HTML]{ffdcb2}0.8538 &\cellcolor[HTML]{FFFFC7}0.2984 &26.00 &\cellcolor[HTML]{FFFFC7}25.24 &\cellcolor[HTML]{FFCCC9}0.8054 &\cellcolor[HTML]{ffdcb2}0.7713 &\cellcolor[HTML]{ffdcb2}0.3480 	 \\
			\thickhline
		\end{tabular}
	}
	\resizebox{\linewidth}{!}{
		\begin{tabular}
			{l|ccccc|ccccc|ccccc}
			\thickhline
			&\multicolumn{5}{c|}{\textit{Rocket}} &\multicolumn{5}{c|}{\textit{Temple}} &\multicolumn{5}{c}{\textit{Yongsil}}  \\
			&PSNR 	&PSNR$^\text{WS}$	&SSIM				&SSIM$^\text{WS}$	&LPIPS
			&PSNR 	&PSNR$^\text{WS}$	&SSIM				&SSIM$^\text{WS}$	&LPIPS
			&PSNR 	&PSNR$^\text{WS}$	&SSIM				&SSIM$^\text{WS}$	&LPIPS\\
			\hline
			Mip-NeRF360~\cite{barron2022mipnerf360}		&\cellcolor[HTML]{FFFFC7}27.92 &\cellcolor[HTML]{FFCCC9}27.47 &0.8531 &\cellcolor[HTML]{FFCCC9}0.8407 &0.3457 &\cellcolor[HTML]{FFFFC7}27.19 &27.41 &0.7458 &0.7554 &0.3825 &\cellcolor[HTML]{FFCCC9}25.57 &\cellcolor[HTML]{FFFFC7}25.55 &\cellcolor[HTML]{FFCCC9}0.7531 &\cellcolor[HTML]{FFFFC7}0.7522 &\cellcolor[HTML]{FFFFC7}0.3393 	 \\
			EgoNeRF~\cite{Choi_2023_CVPR}				&26.82 &26.07 &0.8085 &0.7857 &0.4393 &26.05 &26.04 &0.6930 &0.6961 &0.4699 &24.14 &24.21 &0.6704 &0.6781 &0.4310 	 \\
			LocalRF~\cite{meuleman2023localrf} w/ pose	&27.14 &26.61 &0.8301 &0.8171 &0.3959 &27.08 &\cellcolor[HTML]{FFFFC7}27.45 &\cellcolor[HTML]{FFCCC9}0.7559 &\cellcolor[HTML]{FFCCC9}0.7724 &\cellcolor[HTML]{FFCCC9}0.3537 &24.56 &24.76 &0.7159 &0.7279 &0.3475 	 \\
			D$^2$NeRF~\cite{wu2022d2nerf}				&25.17 &24.22 &0.7576 &0.7225 &0.4560 &23.78 &22.63 &0.6398 &0.6108 &0.5370 &23.81 &23.39 &0.6548 &0.6495 &0.4456 	 \\
			RobustNeRF~\cite{sabour2023robustnerf}		&22.05 &20.89 &0.7633 &0.7214 &0.4741 &22.38 &21.52 &0.6456 &0.6259 &0.5113 &20.15 &19.25 &0.6091 &0.5953 &0.4768 	 \\
			Ours w/ pose								&\cellcolor[HTML]{ffdcb2}28.35 &\cellcolor[HTML]{FFFFC7}27.38 &\cellcolor[HTML]{ffdcb2}0.8534 &0.8359 &\cellcolor[HTML]{FFFFC7}0.3360 &\cellcolor[HTML]{FFCCC9}27.66 &\cellcolor[HTML]{FFCCC9}27.63 &\cellcolor[HTML]{ffdcb2}0.7549 &\cellcolor[HTML]{ffdcb2}0.7663 &\cellcolor[HTML]{ffdcb2}0.3603 &\cellcolor[HTML]{ffdcb2}25.56 &\cellcolor[HTML]{ffdcb2}25.70 &\cellcolor[HTML]{ffdcb2}0.7430 &\cellcolor[HTML]{FFCCC9}0.7566 &\cellcolor[HTML]{FFCCC9}0.3336 	 \\
			\hline
			LocalRF~\cite{meuleman2023localrf} wo/ pose	&27.22 &26.88 &\cellcolor[HTML]{ffdcb2}0.8534 &\cellcolor[HTML]{ffdcb2}0.8396 &\cellcolor[HTML]{FFCCC9}0.3320 &26.55 &26.82 &0.7286 &0.7462 &0.3937 &24.72 &25.11 &0.7185 &0.7418 &0.3433 	 \\
			Ours wo/ pose								&\cellcolor[HTML]{FFCCC9}28.38 &\cellcolor[HTML]{ffdcb2}27.42 &\cellcolor[HTML]{FFCCC9}0.8544 &\cellcolor[HTML]{FFFFC7}0.8369 &\cellcolor[HTML]{ffdcb2}0.3328 &\cellcolor[HTML]{ffdcb2}27.57 &\cellcolor[HTML]{ffdcb2}27.52 &\cellcolor[HTML]{FFFFC7}0.7494 &\cellcolor[HTML]{FFFFC7}0.7600 &\cellcolor[HTML]{FFFFC7}0.3653 &\cellcolor[HTML]{FFFFC7}25.53 &\cellcolor[HTML]{FFCCC9}25.72 &\cellcolor[HTML]{FFFFC7}0.7387 &\cellcolor[HTML]{ffdcb2}0.7540 &\cellcolor[HTML]{ffdcb2}0.3359 	 \\
			\thickhline
		\end{tabular}
	}
	\vspace{3mm}
	\scriptsize
	\centering
	\caption{\label{tb:quantative_syn} 
		Quantitative comparisons of the synthetic video dataset where dynamic objects are inserted. Refer to Figure \ref{fig:comparison_qual_pav} for qualitative comparisons.}	
	\vspace{-3mm}
	\resizebox{\linewidth}{!}{
		\begin{tabular}
			{l|ccccc|ccccc|ccccc}
			\thickhline
			&\multicolumn{5}{c|}{\textit{Sponza}} &\multicolumn{5}{c|}{\textit{Pavillion}} &\multicolumn{5}{c}{\textit{Lone monk}}  \\
			&PSNR 	&PSNR$^\text{WS}$	&SSIM				&SSIM$^\text{WS}$	&LPIPS
			&PSNR 	&PSNR$^\text{WS}$	&SSIM				&SSIM$^\text{WS}$	&LPIPS
			&PSNR 	&PSNR$^\text{WS}$	&SSIM				&SSIM$^\text{WS}$	&LPIPS\\
			\hline
			Mip-NeRF360~\cite{barron2022mipnerf360}, \textit{static only}	&30.91 &32.90 &0.9182 &0.9183 &0.2252 &31.01 &30.12 &0.8810 &0.8736 &0.2218 &26.52 &26.52 &0.7884 &0.7966 &0.3628 	 \\
			Ours wo/ pose, \textit{static only}		&34.22 &34.28 &0.9339 &0.8681 &0.2561 &29.76 &28.58 &0.8532 &0.8360 &0.2771 &26.90 &27.85 &0.8167 &0.8397 &0.3108 	 \\
			\hline
			Mip-NeRF360~\cite{barron2022mipnerf360}		 &22.01 &22.71 &\cellcolor[HTML]{FFFFC7}0.8767 &0.8464 &0.2973 &\cellcolor[HTML]{FFCCC9}30.35 &\cellcolor[HTML]{FFCCC9}29.32 &\cellcolor[HTML]{FFCCC9}0.8798 &\cellcolor[HTML]{FFCCC9}0.8715 &\cellcolor[HTML]{FFCCC9}0.2232 &24.40 &23.94 &0.7799 &0.7858 &0.3751	 \\
			EgoNeRF~\cite{Choi_2023_CVPR}				 &22.14 &22.62 &0.8452 &0.8379 &0.3775 &26.26 &25.48 &0.8146 &0.7983 &0.3736 &23.57 &22.90 &0.7534 &0.7491 &0.4336 	 \\
			LocalRF~\cite{meuleman2023localrf} w/ pose	 &\cellcolor[HTML]{FFFFC7}22.39 &\cellcolor[HTML]{FFFFC7}22.87 &0.8753 &\cellcolor[HTML]{FFFFC7}0.8665 &\cellcolor[HTML]{FFFFC7}0.2725 &27.19 &26.35 &\cellcolor[HTML]{FFFFC7}0.8460 &0.8264 &0.3115 &\cellcolor[HTML]{FFFFC7}25.67 &\cellcolor[HTML]{FFFFC7}25.66 &\cellcolor[HTML]{FFFFC7}0.8074 &\cellcolor[HTML]{FFFFC7}0.8282 &\cellcolor[HTML]{FFFFC7}0.3213 	 \\
			D$^2$NeRF~\cite{wu2022d2nerf}				 &18.05 &18.72 &0.5792 &0.5677 &0.6481 &23.38 &22.51 &0.6927 &0.6646 &0.5912 &18.31 &17.05 &0.5917 &0.5463 &0.6501 	 \\
			RobustNeRF~\cite{sabour2023robustnerf}		 &18.88 &18.86 &0.7242 &0.7029 &0.5489 &24.17 &23.07 &0.7706 &0.7484 &0.4123 &18.73 &17.43 &0.7031 &0.6774 &0.4589 	 \\
			Ours w/ pose								&\cellcolor[HTML]{ffdcb2}33.00 &\cellcolor[HTML]{ffdcb2}32.91 &\cellcolor[HTML]{ffdcb2}0.9232 &\cellcolor[HTML]{ffdcb2}0.9169 &\cellcolor[HTML]{ffdcb2}0.2008 &\cellcolor[HTML]{FFFFC7}29.37 &\cellcolor[HTML]{FFFFC7}28.21 &0.8444 &0.8264 &0.2845 &\cellcolor[HTML]{ffdcb2}26.29 &\cellcolor[HTML]{ffdcb2}27.01 &\cellcolor[HTML]{ffdcb2}0.8093 &\cellcolor[HTML]{ffdcb2}0.8292 &\cellcolor[HTML]{ffdcb2}0.3194 	 \\
			\hline
			LocalRF~\cite{meuleman2023localrf} wo/ pose	 &22.33 &22.76 &0.8710 &0.8603 &0.2781 &28.02 &27.07 &0.8447 &\cellcolor[HTML]{FFFFC7}0.8266 &\cellcolor[HTML]{ffdcb2}0.2758 &25.31 &25.23 &0.8011 &0.8193 &0.3309 \\
			Ours wo/ pose								&\cellcolor[HTML]{FFCCC9}33.35 &\cellcolor[HTML]{FFCCC9}33.32 &\cellcolor[HTML]{FFCCC9}0.9270 &\cellcolor[HTML]{FFCCC9}0.9210 &\cellcolor[HTML]{FFCCC9}0.1925 &\cellcolor[HTML]{ffdcb2}29.71 &\cellcolor[HTML]{ffdcb2}28.55 &\cellcolor[HTML]{ffdcb2}0.8525 &\cellcolor[HTML]{ffdcb2}0.8357 &\cellcolor[HTML]{FFFFC7}0.2767 &\cellcolor[HTML]{FFCCC9}26.74 &\cellcolor[HTML]{FFCCC9}27.68 &\cellcolor[HTML]{FFCCC9}0.8149 &\cellcolor[HTML]{FFCCC9}0.8377 &\cellcolor[HTML]{FFCCC9}0.3140 	 \\
			\thickhline
		\end{tabular}
	}
	\vspace{3mm}
\end{table*}

\begin{figure*}[hptb]
	\centering
	\includegraphics[width=0.97\textwidth]{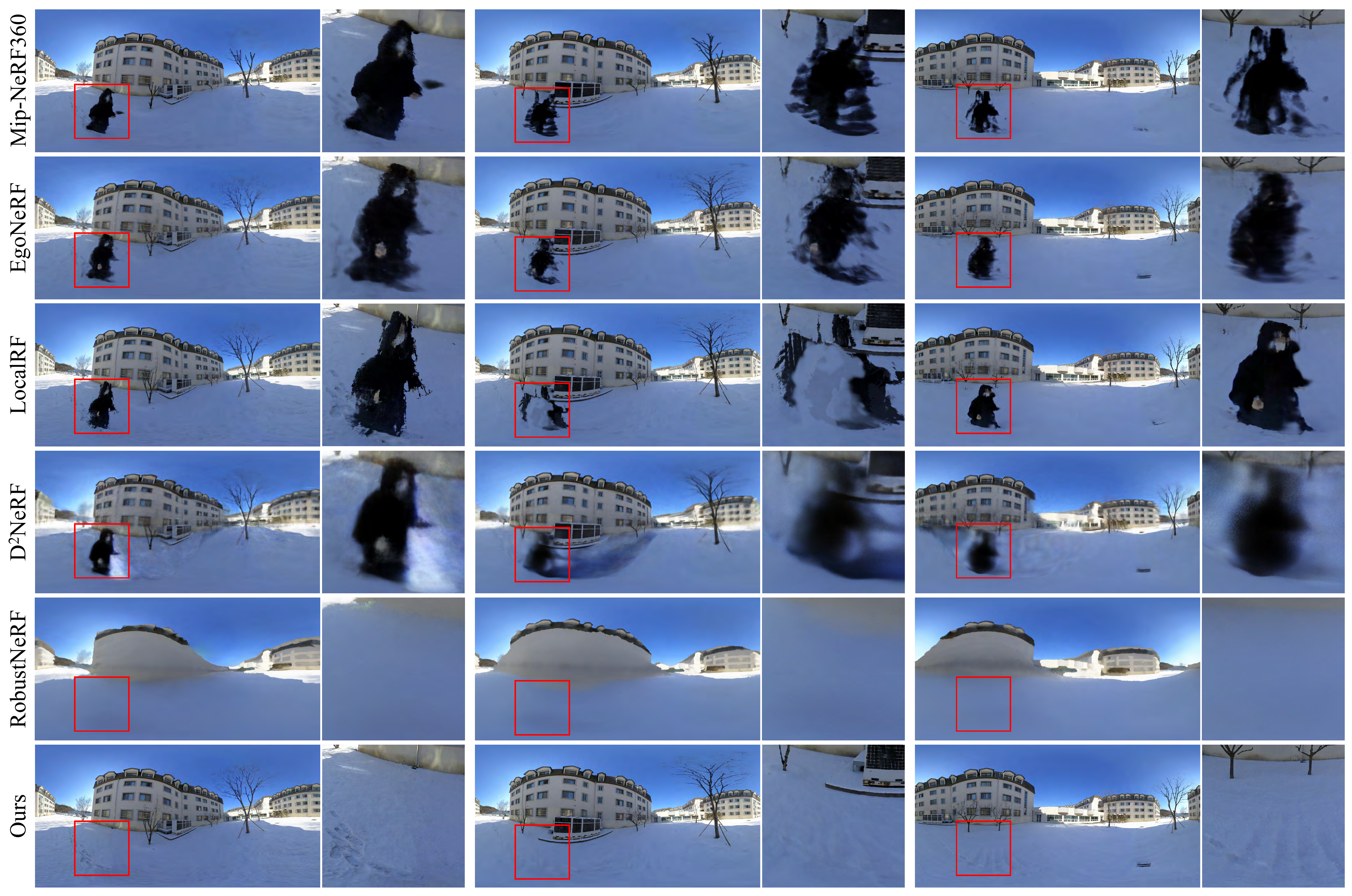}	\\
	\vspace{1mm}
	\includegraphics[width=0.97\textwidth]{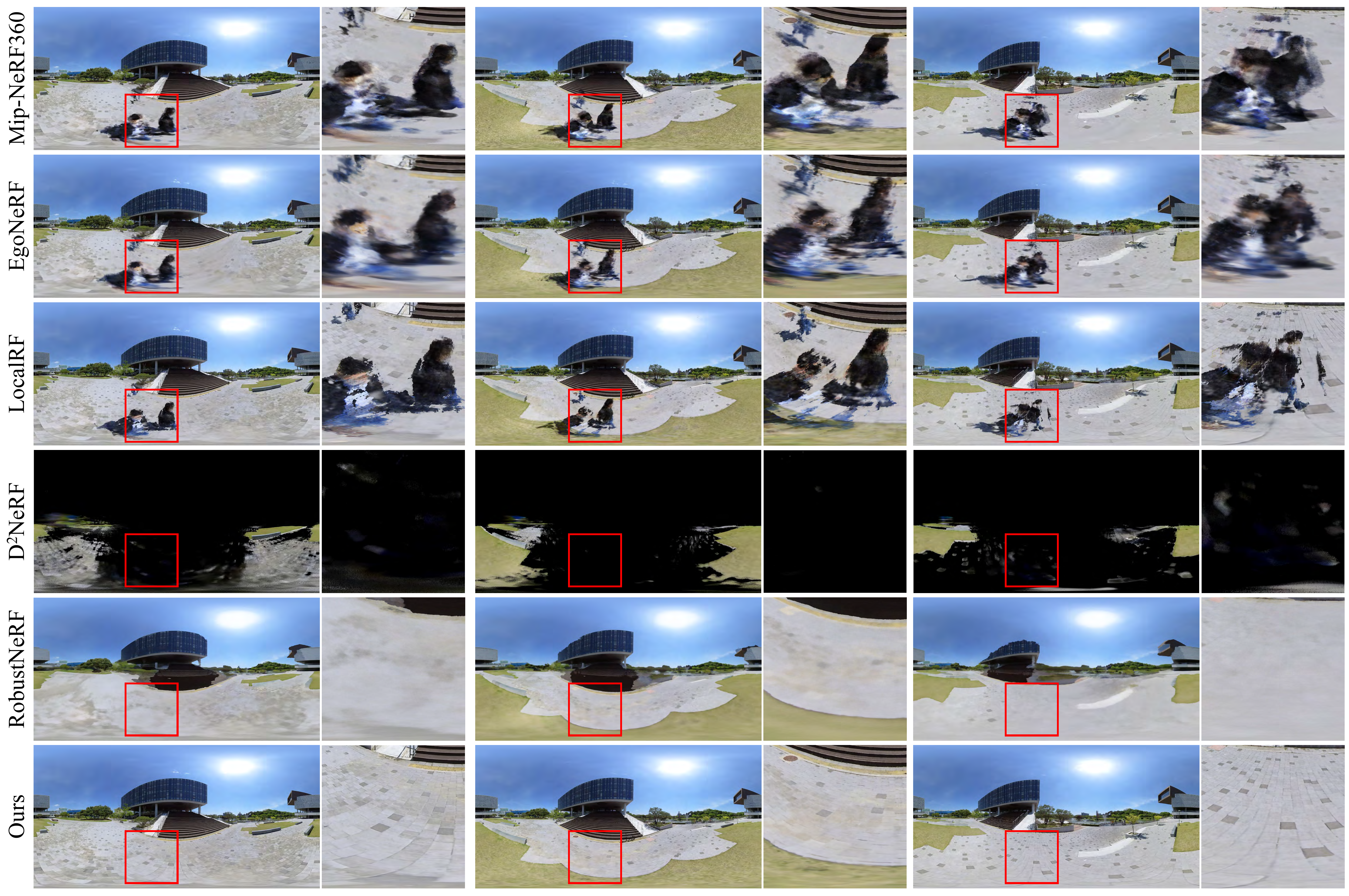}	
	\vspace{-2mm}
	\caption{\label{fig:comparison_qual_dormitory}%
	Qualitative comparisons of the \textit{Dormitory} and \textit{Library} scenes in the real dataset}
\end{figure*}

\begin{figure*}[hptb]
	\centering
	\includegraphics[width=0.97\textwidth]{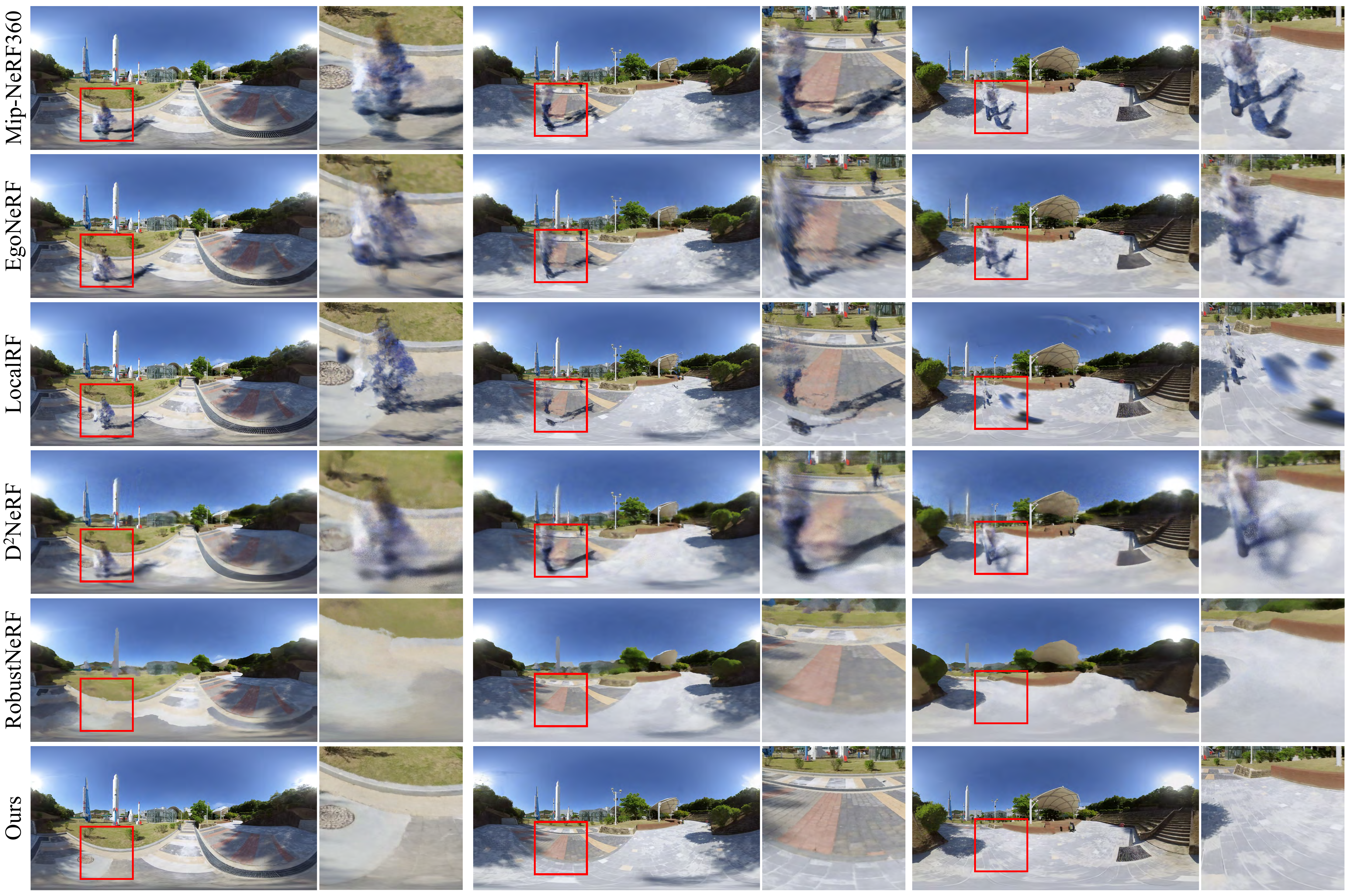}\\
	\vspace{1mm}
	\includegraphics[width=0.97\textwidth]{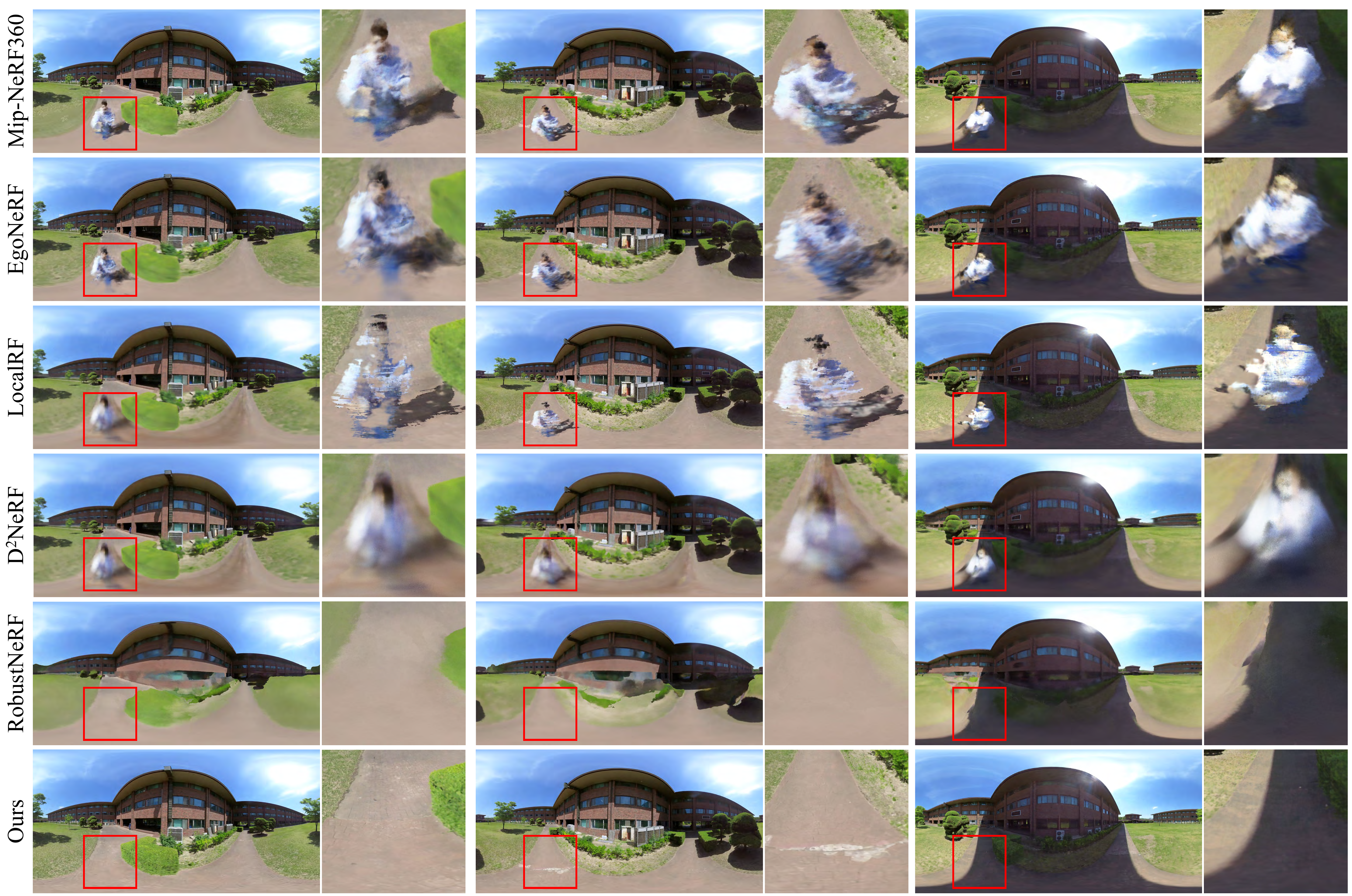}\\
	\vspace{-2mm}
	\caption{\label{fig:comparison_qual_rocket}%
Qualitative comparisons of the \textit{Rocket} and \textit{Red Building} scenes in the real dataset}
	\end{figure*}

\begin{figure*}[hptb]
	\centering
	\includegraphics[width=0.97\textwidth]{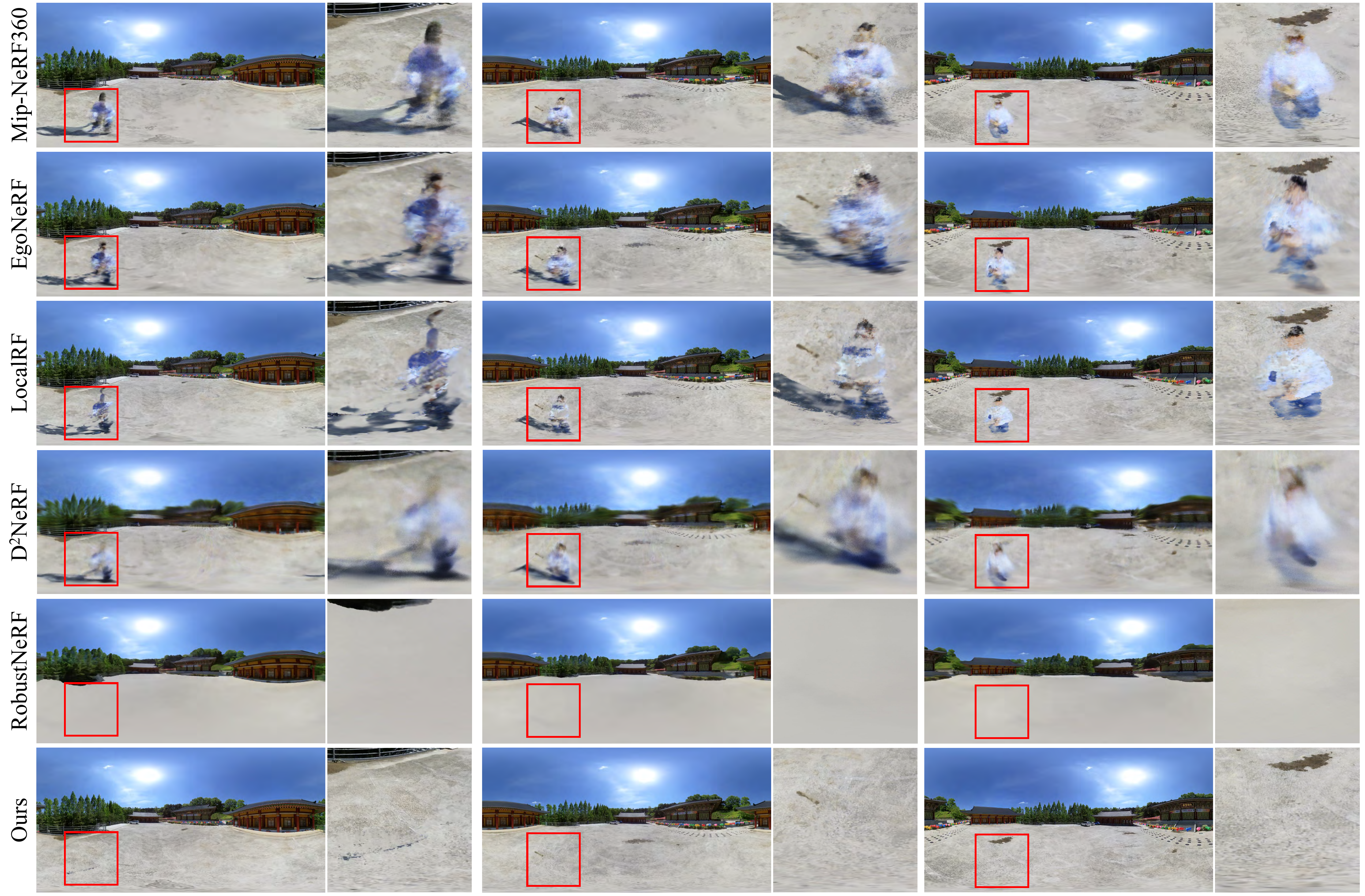}\\
	\vspace{1mm}
	\includegraphics[width=0.97\textwidth]{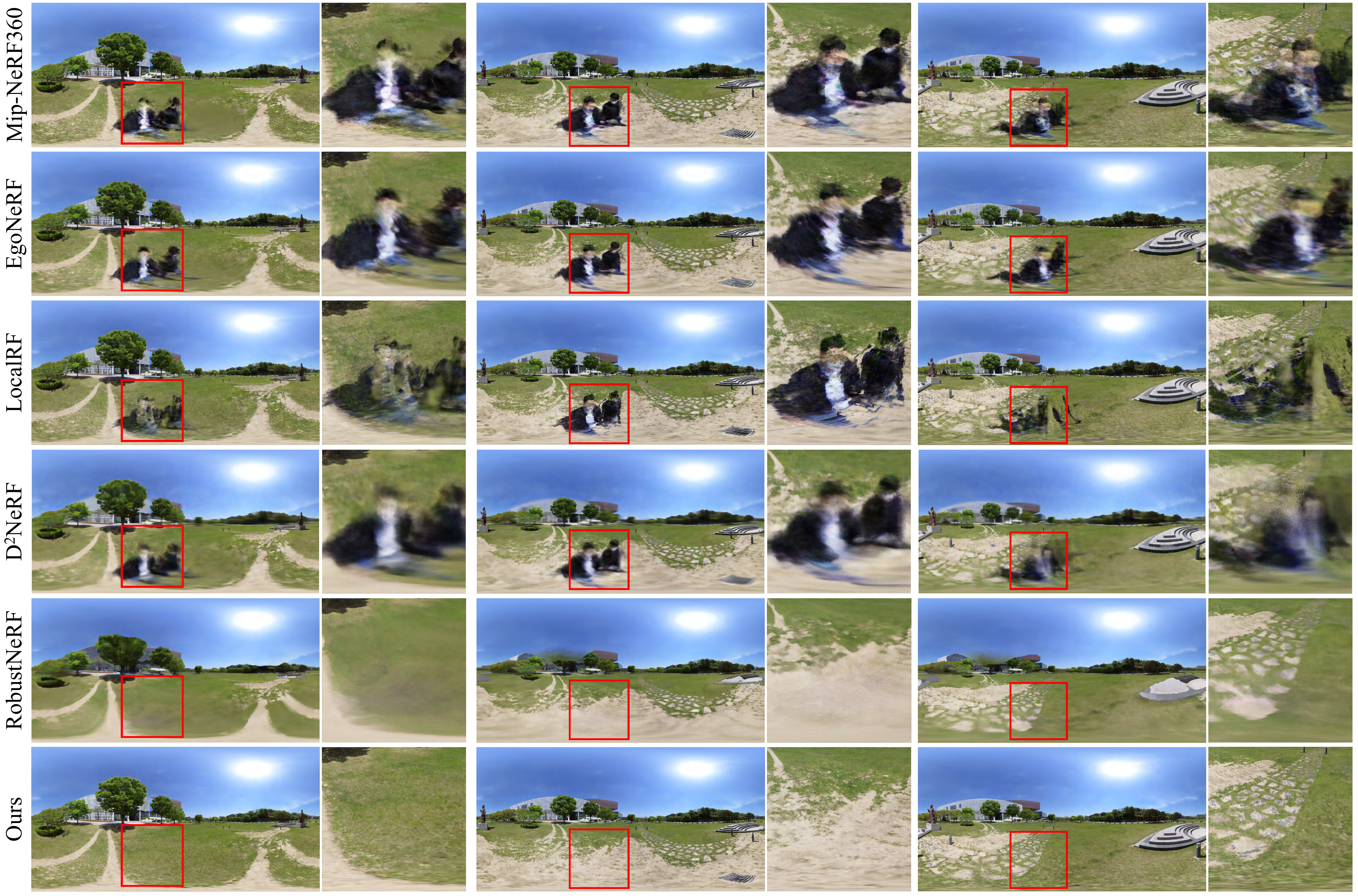}\\
	\vspace{-2mm}	
	\caption{\label{fig:comparison_qual_temple}%
	Qualitative comparisons of the \textit{Temple} and \textit{Yongsil} scenes in the real dataset}
\end{figure*}

\begin{figure*}[hptb]
	\vspace{1mm}
	\centering
	\includegraphics[width=1.0\textwidth]{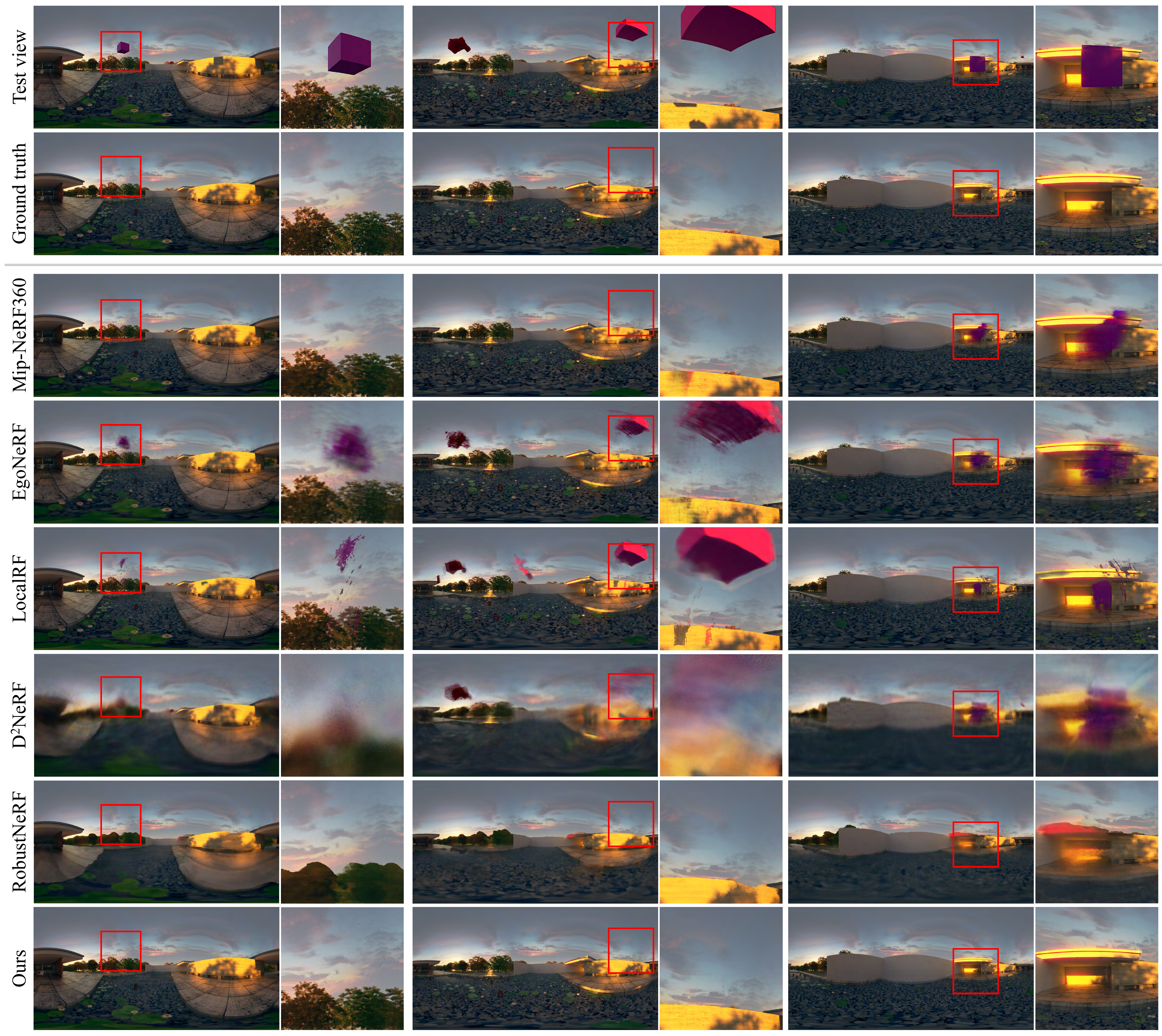}\\
	\caption{\label{fig:comparison_qual_pav}%
	Qualitative comparisons of the \textit{Pavillion} scene in the synthetic video dataset where synthetic moving objects are inserted. 
	}
\end{figure*}

\subsection{Hyperparameters}
We set the volumetric density channel to $8$ for the RF block's volumetric density channel and $24$ for its color channel. 
For both the MLP used to decode color from radiance blocks and the MLP used to estimate dynamic components, we use $2$ layers with $128$ hidden nodes. 
We adopt the Adam optimizer~\cite{KingBa15} with $\beta_1=0.9$ and $\beta_2=0.99$ to train all parameters.
To account for the distinct convergence speeds of each parameter, we use different learning rate schedules. 
The learning rate for radiance blocks and neural feature planes starts at $2\times10^{-2}$ and exponentially decays to $2\times10^{-3}$.
The learning rate for the camera matrix starts at
$5\times10^{-3}$ and exponentially decays to $5\times10^{-4}$.
To prevent excessive change in the mask MLP, which is used to estimate motion masks of entire frames, 
we set the learning rate to start at $5\times10^{-4}$ and exponentially decay to $5\times10^{-6}$.

\section{Motion Mask Estimation}
\label{sec:motionmask}
In recent works such as~\cite{martin2021nerf, li2023dynibar}, dynamic objects, which are temporally dependent transient data that are independent of global geometry, are excluded under the Bayesian learning framework.
This is because dynamic components introduce high epistemic uncertainty, which slows down the learning process and results in high errors when performing tasks such as volume-based view synthesis that require 3D reconstruction.
To implicitly address dynamic components, RobustNeRF~\cite{sabour2023robustnerf} down-weights the loss of regions with high uncertainty. 
However, in our approach, we explicitly separate dynamic components from stationary geometry.
In the task of predicting dynamic color based on pixel position input imposed on the mask module, the uncertainty of transient data significantly decreases because of a lack of geometrical constraint. 
Therefore, by performing both tasks simultaneously, the mask module converges faster to predict dynamic color values and alpha values compared to radiance fields.
The segmentation of dynamic components allows radiance fields to ignore them during the training process, focusing solely on static elements.
Figure~\ref{fig:mask_results} shows examples of our mask results.

When training the mask module to estimate motion masks, there is a problem with learning dynamic alpha values because of factorization ambiguity. 
To solve this issue, we supervise the estimated dynamic color by ground truth with added Gaussian noise. This helps to narrow down the possibilities for alpha values, making it easier to obtain a mask closer to a binary value. 
Without this supervision, the motion mask prediction struggles to reach a binary value, which can result in dynamic artifacts on the static geometry. 
Refer to Figure~\ref{fig:mask_wo_rgb} to see an illustration of this issue.

\begin{table}[t]
	\small
	\centering
	\caption{\label{tb:time} 
		Our method performs better than existing techniques when training on the real dataset with slightly increased computing time.
	}
	\vspace{-2mm}
	\resizebox{\linewidth}{!}{
		\begin{tabular}
			{l|ccccc}
			\thickhline		
			& Iteration	& Time $\downarrow$	&PSNR $\uparrow$ & SSIM $\uparrow$ &LPIPS $\downarrow$\\
			\hline
			Mip-NeRF360~\cite{barron2022mipnerf360}		 &$250$K  &20 hrs  &\cellcolor[HTML]{FFFFC7}26.88  &\cellcolor[HTML]{FFFFC7}0.8094  &0.3583 	 \\
			EgoNeRF~\cite{Choi_2023_CVPR}				  &$100$K  &\cellcolor[HTML]{FFCCC9}5.2 hrs  &25.95  &0.7609  &0.4383 	 \\
			LocalRF~\cite{meuleman2023localrf}	  &$155$K  &\cellcolor[HTML]{FFFFC7}8.5 hrs  &26.56  &0.8034  &\cellcolor[HTML]{FFFFC7}0.3410 	 \\
			D$^2$NeRF~\cite{wu2022d2nerf}				  &$100$K  &\cellcolor[HTML]{ffdcb2}6 hrs  &20.95  &0.6105  &0.5100 	 \\
			RobustNeRF~\cite{sabour2023robustnerf}		  &$500$K  &40 hrs  &20.78  &0.7093  &0.4864 	 \\
			\hline
			Ours	 &$120$K  &8.7 hrs  &\cellcolor[HTML]{ffdcb2}27.61  &\cellcolor[HTML]{ffdcb2}0.8135  &\cellcolor[HTML]{ffdcb2}0.3372 	 \\
			Ours	 &$155$K  &12.5 hrs  &\cellcolor[HTML]{FFCCC9}27.73  &\cellcolor[HTML]{FFCCC9}0.8165  &\cellcolor[HTML]{FFCCC9}0.3297 	 \\
			\thickhline
		\end{tabular}
	}
	\vspace{-3mm}
\end{table}

\section{Comparison Details}
\label{sec:compararison_details}
To compare our results, we use the official implementations of D$^2$NeRF~\cite{wu2022d2nerf}, EgoNeRF~\cite{Choi_2023_CVPR}, and LocalRF~\cite{meuleman2023localrf}. 
We employ the Multi-NeRF~\cite{multinerf2022} code for Mip-NeRF360~\cite{barron2022mipnerf360} and RobustNeRF~\cite{sabour2023robustnerf}. 
All experiments are conducted on a machine with a single NVIDIA A6000 GPU and an Intel Xeon Silver 4214R 2.40 GHz CPU with 256 GB RAM. 
The baseline methods use the default hyperparameters, including the number of iterations, except for RobustNeRF, for which we use $500$K iterations due to its slow convergence by the IRLS-based approach. 
We present the training time and metrics in Table~\ref{tb:time}. 
Additionally, we present the results of our method with $120$K iterations, which show comparable training time to LocalRF. 
We also use the results of our method with $155$K iterations, chosen for its ability to capture intricate details.

\section{Additional Results}
\label{sec:additional_results}
\begin{figure*}[hptb]
	\vspace{1mm}
	\centering
	\includegraphics[width=0.8\textwidth]{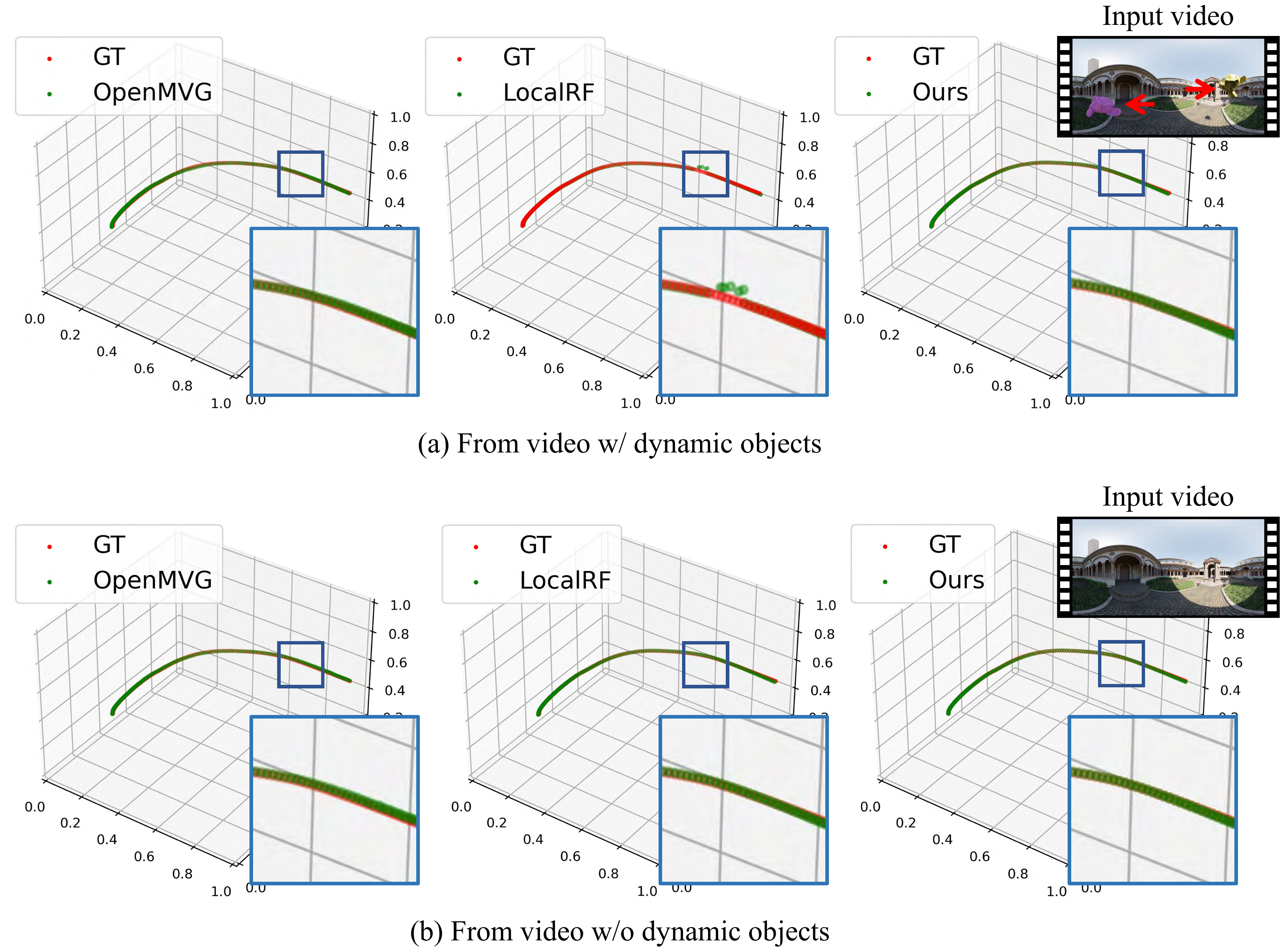}\\
	\caption{\label{fig:supple_poses}%
	In the \textit{Lone Monk} scene, we compare the estimated pose with the ground-truth pose both with and without dynamic objects. 
	The results with dynamic objects are shown in (a), and without dynamic objects in (b). 
	Our method is capable of robustly estimating frame poses, regardless of the presence of dynamic objects.}
\end{figure*}

We present additional results for the real dataset. 
Figures~\ref{fig:comparison_qual_dormitory},~\ref{fig:comparison_qual_rocket}, and~\ref{fig:comparison_qual_temple} display the results, and the corresponding metrics are available in Table~\ref{tb:quantative_real}. 
Additionally, we include the results from the \textit{Pavillion} scene in the synthetic dataset, which is presented in Figure~\ref{fig:comparison_qual_pav}. 
The respective metrics are available in Table~\ref{tb:quantative_syn}. 
We also include the pose estimation results in Figure~\ref{fig:supple_poses} and Table~\ref{tb:pose}.

\begin{table*}[htpb]
	\small
	\centering
	\caption{\label{tb:pose} 
		Quantitative comparisons of the pose accuracy on the synthetic dataset. Our method shows superior performance, especially in scenarios where dynamic objects are present.
		Refer to Figure~\ref{fig:supple_poses} for qualitative comparisons.
	}
	\vspace{-1mm}
	\resizebox{0.8\linewidth}{!}{
	\begin{tabular}
		{l|ccc|ccc|ccc}
		\thickhline		
		& \multicolumn{3}{c|}{\textit{Sponza}}&\multicolumn{3}{c|}{\textit{Pavillion}} &\multicolumn{3}{c}{\textit{Lone monk}}\\
		& RPE$_r$ $\downarrow$ 	& RPE$_t$ $\downarrow$	&ATE $\downarrow$ & RPE$_r$ $\downarrow$	& RPE$_t$ $\downarrow$	&ATE $\downarrow$ & RPE$_r$ $\downarrow$	& RPE$_t$ $\downarrow$	&ATE  $\downarrow$\\
		\hline
		OpenMVG	&\cellcolor[HTML]{ffdcb2}0.09310	&\cellcolor[HTML]{ffdcb2}0.00045 	&\cellcolor[HTML]{FFCCC9}{0.00032}	&\cellcolor[HTML]{ffdcb2}0.17250	&\cellcolor[HTML]{FFFFC7}0.01962 	&\cellcolor[HTML]{ffdcb2}0.00347 &\cellcolor[HTML]{FFFFC7}0.05726	&\cellcolor[HTML]{FFFFC7}0.03388	&\cellcolor[HTML]{ffdcb2}0.00263 	\\
		LocalRF &\cellcolor[HTML]{FFCCC9}{0.09309}	&\cellcolor[HTML]{FFFFC7}0.00074 	&\cellcolor[HTML]{FFFFC7}0.00336	&\cellcolor[HTML]{FFCCC9}{0.17133}	&\cellcolor[HTML]{ffdcb2}0.00065 	&\cellcolor[HTML]{FFFFC7}0.00417 &\cellcolor[HTML]{ffdcb2}0.04770	&\cellcolor[HTML]{ffdcb2}0.00145	&\cellcolor[HTML]{FFFFC7}0.00374 	\\
		Ours	&\cellcolor[HTML]{FFCCC9}{0.09309}	&\cellcolor[HTML]{FFCCC9}{0.00020} 	&\cellcolor[HTML]{ffdcb2}0.00207	&\cellcolor[HTML]{FFCCC9}{0.17133}	&\cellcolor[HTML]{FFCCC9}{0.00056} 	&\cellcolor[HTML]{FFCCC9}{0.00116} &\cellcolor[HTML]{FFCCC9}{0.04753}	&\cellcolor[HTML]{FFCCC9}{0.00143}	&\cellcolor[HTML]{FFCCC9}{0.00236} 	\\
		\hline
		OpenMVG, \textit{static only} &\cellcolor[HTML]{FFCCC9}{0.09309}	&\cellcolor[HTML]{FFFFC7}0.00047 	&\cellcolor[HTML]{FFCCC9}{0.00036}	&\cellcolor[HTML]{ffdcb2}0.17255	&\cellcolor[HTML]{FFFFC7}0.01957 	&\cellcolor[HTML]{FFFFC7}0.00332 &\cellcolor[HTML]{FFFFC7}0.05552	&\cellcolor[HTML]{FFFFC7}0.03384	&\cellcolor[HTML]{ffdcb2}0.00195 	\\
		LocalRF, \textit{static only}	&\cellcolor[HTML]{FFCCC9}{0.09309}	&\cellcolor[HTML]{FFCCC9}{0.00018} 	&\cellcolor[HTML]{ffdcb2}0.00173	&\cellcolor[HTML]{FFCCC9}{0.17133}	&\cellcolor[HTML]{FFCCC9}{0.00052} 	&\cellcolor[HTML]{ffdcb2}0.00222 &\cellcolor[HTML]{FFCCC9}{0.04753}	&\cellcolor[HTML]{FFCCC9}{0.00143}	&\cellcolor[HTML]{FFFFC7}0.00230 	\\
		Ours, \textit{static only}	&\cellcolor[HTML]{FFCCC9}{0.09309}	&\cellcolor[HTML]{ffdcb2}0.00020 	&\cellcolor[HTML]{FFFFC7}0.00207	&\cellcolor[HTML]{FFCCC9}{0.17133}	&\cellcolor[HTML]{ffdcb2}0.00058 	&\cellcolor[HTML]{FFCCC9}{0.00135} &\cellcolor[HTML]{FFCCC9}{0.04753}	&\cellcolor[HTML]{ffdcb2}0.00146	&\cellcolor[HTML]{FFCCC9}{0.00153} 	\\
		\thickhline
	\end{tabular}
	}
	\vspace{-3mm}
\end{table*}

{
    \small
    \bibliographystyle{ieeenat_fullname}
    \bibliography{OmniLocalRF-reference}

\begin{thebibliography}{59}
\providecommand{\natexlab}[1]{#1}
\providecommand{\url}[1]{\texttt{#1}}
\expandafter\ifx\csname urlstyle\endcsname\relax
  \providecommand{\doi}[1]{doi: #1}\else
  \providecommand{\doi}{doi: \begingroup \urlstyle{rm}\Url}\fi

\bibitem[Attal et~al.(2020)Attal, Ling, Gokaslan, Richardt, and
  Tompkin]{Attal20ECCV}
Benjamin Attal, Selena Ling, Aaron Gokaslan, Christian Richardt, and James
  Tompkin.
\newblock {{MatryODShka}: Real-Time {6DoF} Video View Synthesis using
  Multi-Sphere Images}.
\newblock In \emph{ECCV}, 2020.

\bibitem[Barron et~al.(2022)Barron, Mildenhall, Verbin, Srinivasan, and
  Hedman]{barron2022mipnerf360}
Jonathan~T Barron, Ben Mildenhall, Dor Verbin, Pratul~P Srinivasan, and Peter
  Hedman.
\newblock {Mip-NeRF 360: Unbounded Anti-Aliased Neural Radiance Fields}.
\newblock In \emph{CVPR}, pages 5470--5479, 2022.

\bibitem[Bertel et~al.(2020)Bertel, Yuan, Lindroos, and Richardt]{OmniPhotos}
Tobias Bertel, Mingze Yuan, Reuben Lindroos, and Christian Richardt.
\newblock {{OmniPhotos}: Casual 360° {VR} Photography}.
\newblock \emph{TOG}, 39\penalty0 (6):\penalty0 266:1--12, 2020.

\bibitem[Bian et~al.(2023)Bian, Wang, Li, Bian, and Prisacariu]{bian2023nope}
Wenjing Bian, Zirui Wang, Kejie Li, Jia-Wang Bian, and Victor~Adrian
  Prisacariu.
\newblock {NoPe-NeRF: Optimising Neural Radiance Field with No Pose Prior}.
\newblock In \emph{CVPR}, pages 4160--4169, 2023.

\bibitem[Broxton et~al.(2019)Broxton, Busch, Dourgarian, DuVall, Erickson,
  Evangelakos, Flynn, Overbeck, Whalen, and Debevec]{broxton19lowcost}
Michael Broxton, Jay Busch, Jason Dourgarian, Matthew DuVall, Daniel Erickson,
  Dan Evangelakos, John Flynn, Ryan Overbeck, Matt Whalen, and Paul Debevec.
\newblock {A Low Cost Multi-Camera Array for Panoramic Light Field Video
  Capture}.
\newblock In \emph{SIGGRAPH Asia Posters}, New York, NY, USA, 2019. Association
  for Computing Machinery.

\bibitem[Broxton et~al.(2020{\natexlab{a}})Broxton, Busch, Dourgarian, DuVall,
  Erickson, Evangelakos, Flynn, Hedman, Overbeck, Whalen, and
  Debevec]{broxton2020deepview}
Michael Broxton, Jay Busch, Jason Dourgarian, Matthew DuVall, Daniel Erickson,
  Dan Evangelakos, John Flynn, Peter Hedman, Ryan Overbeck, Matt Whalen, and
  Paul Debevec.
\newblock {DeepView Immersive Light Field Video}.
\newblock In \emph{ACM SIGGRAPH Immersive Pavilion}. Association for Computing
  Machinery, 2020{\natexlab{a}}.

\bibitem[Broxton et~al.(2020{\natexlab{b}})Broxton, Flynn, Overbeck, Erickson,
  Hedman, Duvall, Dourgarian, Busch, Whalen, and Debevec]{broxton2020immersive}
Michael Broxton, John Flynn, Ryan Overbeck, Daniel Erickson, Peter Hedman,
  Matthew Duvall, Jason Dourgarian, Jay Busch, Matt Whalen, and Paul Debevec.
\newblock {Immersive Light Field Video with a Layered Mesh Representation}.
\newblock \emph{TOG}, 39\penalty0 (4), 2020{\natexlab{b}}.

\bibitem[Chen et~al.(2022)Chen, Xu, Geiger, Yu, and Su]{chen2022tensorf}
Anpei Chen, Zexiang Xu, Andreas Geiger, Jingyi Yu, and Hao Su.
\newblock {TensoRF: Tensorial Radiance Fields}.
\newblock In \emph{ECCV}, pages 333--350. Springer, 2022.

\bibitem[Chen et~al.(2018)Chen, Zhu, Papandreou, Schroff, and
  Adam]{chen2018encoder}
Liang-Chieh Chen, Yukun Zhu, George Papandreou, Florian Schroff, and Hartwig
  Adam.
\newblock {Encoder-Decoder with Atrous Separable Convolution for Semantic Image
  Segmentation}.
\newblock In \emph{ECCV}, pages 801--818, 2018.

\bibitem[Cheng et~al.(2020)Cheng, Collins, Zhu, Liu, Huang, Adam, and
  Chen]{cheng2020panoptic}
Bowen Cheng, Maxwell~D Collins, Yukun Zhu, Ting Liu, Thomas~S Huang, Hartwig
  Adam, and Liang-Chieh Chen.
\newblock {Panoptic-Deeplab: A Simple, Strong, and Fast Baseline for Bottom-Up
  Panoptic Segmentation}.
\newblock In \emph{CVPR}, pages 12475--12485, 2020.

\bibitem[Choi et~al.(2023)Choi, Kim, and Kim]{Choi_2023_CVPR}
Changwoon Choi, Sang~Min Kim, and Young~Min Kim.
\newblock {Balanced Spherical Grid for Egocentric View Synthesis}.
\newblock In \emph{CVPR}, pages 16590--16599, 2023.

\bibitem[Fridovich-Keil et~al.(2023)Fridovich-Keil, Meanti, Warburg, Recht, and
  Kanazawa]{fridovich2023k}
Sara Fridovich-Keil, Giacomo Meanti, Frederik~Rahb{\ae}k Warburg, Benjamin
  Recht, and Angjoo Kanazawa.
\newblock {K-planes: Explicit Radiance Fields in Space, Time, and Appearance}.
\newblock In \emph{CVPR}, pages 12479--12488, 2023.

\bibitem[Granados et~al.(2008)Granados, Seidel, and
  Lensch]{granados2008background}
Miguel Granados, Hans-Peter Seidel, and Hendrik~PA Lensch.
\newblock {Background Estimation from Non-Time Sequence Images}.
\newblock In \emph{Proceedings of Graphics Interface 2008}, pages 33--40, 2008.

\bibitem[Granados et~al.(2012)Granados, Tompkin, Kim, Grau, Kautz, and
  Theobalt]{granados2012not}
Miguel Granados, James Tompkin, Kwang~In Kim, Oliver Grau, Jan Kautz, and
  Christian Theobalt.
\newblock {How Not to Be Seen—Object Removal from Videos of Crowded Scenes}.
\newblock In \emph{CGF}, pages 219--228. Wiley Online Library, 2012.

\bibitem[Guyon et~al.(2012{\natexlab{a}})Guyon, Bouwmans, and
  Zahzah]{guyon2012foreground}
Charles Guyon, Thierry Bouwmans, and El-Hadi Zahzah.
\newblock {Foreground Detection via Robust Low Rank Matrix Decomposition
  Including Spatio-Temporal Constraint}.
\newblock In \emph{ACCV}, pages 315--320. Springer, 2012{\natexlab{a}}.

\bibitem[Guyon et~al.(2012{\natexlab{b}})Guyon, Bouwmans, and
  Zahzah]{guyon2012moving}
Charles Guyon, Thierry Bouwmans, and El-Hadi Zahzah.
\newblock {Moving Object Detection via Robust Low Rank Matrix Decomposition
  with IRLS Scheme}.
\newblock In \emph{Int. Symp. Visual Computing}, pages 665--674. Springer,
  2012{\natexlab{b}}.

\bibitem[He et~al.(2017)He, Gkioxari, Doll{\'a}r, and Girshick]{he2017mask}
Kaiming He, Georgia Gkioxari, Piotr Doll{\'a}r, and Ross Girshick.
\newblock {Mask R-CNN}.
\newblock In \emph{ICCV}, pages 2961--2969, 2017.

\bibitem[Jang et~al.(2022)Jang, Meuleman, Kang, Kim, Richardt, and
  Kim]{Egocentric}
Hyeonjoong Jang, Andréas Meuleman, Dahyun Kang, Donggun Kim, Christian
  Richardt, and Min~H. Kim.
\newblock {Egocentric Scene Reconstruction from an Omnidirectional Video}.
\newblock \emph{TOG}, 41\penalty0 (4), 2022.

\bibitem[Jeong et~al.(2021)Jeong, Ahn, Choy, Anandkumar, Cho, and
  Park]{jeong2021self}
Yoonwoo Jeong, Seokjun Ahn, Christopher Choy, Anima Anandkumar, Minsu Cho, and
  Jaesik Park.
\newblock {Self-Calibrating Neural Radiance Fields}.
\newblock In \emph{ICCV}, pages 5846--5854, 2021.

\bibitem[Kopf et~al.(2021)Kopf, Rong, and Huang]{kopf2021robust}
Johannes Kopf, Xuejian Rong, and Jia-Bin Huang.
\newblock {Robust Consistent Video Depth Estimation}.
\newblock In \emph{CVPR}, pages 1611--1621, 2021.

\bibitem[Li et~al.(2021)Li, Niklaus, Snavely, and Wang]{li2021neural}
Zhengqi Li, Simon Niklaus, Noah Snavely, and Oliver Wang.
\newblock {Neural Scene Flow Fields for Space-Time View Synthesis of Dynamic
  Scenes}.
\newblock In \emph{CVPR}, 2021.

\bibitem[Li et~al.(2023)Li, Wang, Cole, Tucker, and Snavely]{li2023dynibar}
Zhengqi Li, Qianqian Wang, Forrester Cole, Richard Tucker, and Noah Snavely.
\newblock {DynIBaR: Neural Dynamic Image-Based Rendering}.
\newblock In \emph{CVPR}, 2023.

\bibitem[Lin et~al.(2021)Lin, Ma, Torralba, and Lucey]{lin2021barf}
Chen-Hsuan Lin, Wei-Chiu Ma, Antonio Torralba, and Simon Lucey.
\newblock {BARF: Bundle-Adjusting Neural Radiance Fields}.
\newblock In \emph{CVPR}, pages 5741--5751, 2021.

\bibitem[Lin et~al.(2023)Lin, Gao, Huang, Kim, Wang, Zwicker, and
  Saraf]{Lin_2023_ICCV}
Geng Lin, Chen Gao, Jia-Bin Huang, Changil Kim, Yipeng Wang, Matthias Zwicker,
  and Ayush Saraf.
\newblock {OmnimatteRF: Robust Omnimatte with 3D Background Modeling}.
\newblock In \emph{ICCV}, 2023.

\bibitem[Liu et~al.(2023)Liu, Gao, Meuleman, Tseng, Saraf, Kim, Chuang, Kopf,
  and Huang]{liu2023robust}
Yu-Lun Liu, Chen Gao, Andreas Meuleman, Hung-Yu Tseng, Ayush Saraf, Changil
  Kim, Yung-Yu Chuang, Johannes Kopf, and Jia-Bin Huang.
\newblock {Robust Dynamic Radiance Fields}.
\newblock In \emph{CVPR}, 2023.

\bibitem[Meuleman et~al.(2023)Meuleman, Liu, Gao, Huang, Kim, Kim, and
  Kopf]{meuleman2023localrf}
Andreas Meuleman, Yu-Lun Liu, Chen Gao, Jia-Bin Huang, Changil Kim, Min~H Kim,
  and Johannes Kopf.
\newblock {Progressively Optimized Local Radiance Fields for Robust View
  Synthesis}.
\newblock In \emph{CVPR}, pages 16539--16548, 2023.

\bibitem[Mildenhall et~al.(2020)Mildenhall, Srinivasan, Tancik, Barron,
  Ramamoorthi, and Ng]{mildenhall2020nerf}
Ben Mildenhall, Pratul~P Srinivasan, Matthew Tancik, Jonathan~T Barron, Ravi
  Ramamoorthi, and Ren Ng.
\newblock {NeRF: Representing Scenes as Neural Radiance Fields for View
  Synthesis}.
\newblock In \emph{ECCV}, pages 405--421. Springer, 2020.

\bibitem[Mirzaei et~al.(2023)Mirzaei, Aumentado-Armstrong, Derpanis, Kelly,
  Brubaker, Gilitschenski, and Levinshtein]{spinnerf}
Ashkan Mirzaei, Tristan Aumentado-Armstrong, Konstantinos~G Derpanis, Jonathan
  Kelly, Marcus~A Brubaker, Igor Gilitschenski, and Alex Levinshtein.
\newblock {SPIn-NeRF: Multiview Segmentation and Perceptual Inpainting with
  Neural Radiance Fields}.
\newblock In \emph{CVPR}, pages 20669--20679, 2023.

\bibitem[Moulon et~al.(2017)Moulon, Monasse, Perrot, and
  Marlet]{moulon2017openmvg}
Pierre Moulon, Pascal Monasse, Romuald Perrot, and Renaud Marlet.
\newblock {OpenMVG: Open Multiple View Geometry}.
\newblock In \emph{Int. Workshop on Reproducible Research in Pattern
  Recognition}, pages 60--74. Springer, 2017.

\bibitem[M{\"u}ller et~al.(2022)M{\"u}ller, Evans, Schied, and
  Keller]{muller2022instant}
Thomas M{\"u}ller, Alex Evans, Christoph Schied, and Alexander Keller.
\newblock {Instant Neural Graphics Primitives with a Multiresolution Hash
  Encoding}.
\newblock \emph{TOG}, 41\penalty0 (4):\penalty0 1--15, 2022.

\bibitem[Overbeck et~al.(2018)Overbeck, Erickson, Evangelakos, Pharr, and
  Debevec]{overbeck18system}
Ryan~S. Overbeck, Daniel Erickson, Daniel Evangelakos, Matt Pharr, and Paul
  Debevec.
\newblock {A System for Acquiring, Processing, and Rendering Panoramic Light
  Field Stills for Virtual Reality}.
\newblock \emph{TOG}, 37\penalty0 (6), 2018.

\bibitem[Park et~al.(2021{\natexlab{a}})Park, Sinha, Barron, Bouaziz, Goldman,
  Seitz, and Martin-Brualla]{park2021nerfies}
Keunhong Park, Utkarsh Sinha, Jonathan~T Barron, Sofien Bouaziz, Dan~B Goldman,
  Steven~M Seitz, and Ricardo Martin-Brualla.
\newblock {Nerfies: Deformable Neural Radiance Fields}.
\newblock In \emph{ICCV}, pages 5865--5874, 2021{\natexlab{a}}.

\bibitem[Park et~al.(2021{\natexlab{b}})Park, Sinha, Hedman, Barron, Bouaziz,
  Goldman, Martin-Brualla, and Seitz]{park2021hypernerf}
Keunhong Park, Utkarsh Sinha, Peter Hedman, Jonathan~T Barron, Sofien Bouaziz,
  Dan~B Goldman, Ricardo Martin-Brualla, and Steven~M Seitz.
\newblock {HyperNeRF: a Higher-Dimensional Representation for Topologically
  Varying Neural Radiance Fields}.
\newblock \emph{TOG}, 40\penalty0 (6):\penalty0 1--12, 2021{\natexlab{b}}.

\bibitem[Poole et~al.(2022)Poole, Jain, Barron, and
  Mildenhall]{poole2022dreamfusion}
Ben Poole, Ajay Jain, Jonathan~T Barron, and Ben Mildenhall.
\newblock {DreamFusion: Text-to-3D using 2D Diffusion}.
\newblock \emph{arXiv preprint arXiv:2209.14988}, 2022.

\bibitem[Pozo et~al.(2019)Pozo, Toksvig, Schrager, Hsu, Mathur,
  Sorkine-Hornung, Szeliski, and Cabral]{pozo19integrated}
Albert~Parra Pozo, Michael Toksvig, Terry~Filiba Schrager, Joyce Hsu, Uday
  Mathur, Alexander Sorkine-Hornung, Rick Szeliski, and Brian Cabral.
\newblock {An Integrated 6DoF Video Camera and System Design}.
\newblock \emph{TOG}, 38\penalty0 (6), 2019.

\bibitem[Rematas et~al.(2022)Rematas, Liu, Srinivasan, Barron, Tagliasacchi,
  Funkhouser, and Ferrari]{rematas2022urban}
Konstantinos Rematas, Andrew Liu, Pratul~P Srinivasan, Jonathan~T Barron,
  Andrea Tagliasacchi, Thomas Funkhouser, and Vittorio Ferrari.
\newblock {Urban Radiance Fields}.
\newblock In \emph{CVPR}, pages 12932--12942, 2022.

\bibitem[Sabour et~al.(2023)Sabour, Vora, Duckworth, Krasin, Fleet, and
  Tagliasacchi]{sabour2023robustnerf}
Sara Sabour, Suhani Vora, Daniel Duckworth, Ivan Krasin, David~J Fleet, and
  Andrea Tagliasacchi.
\newblock {RobustNeRF: Ignoring Distractors with Robust Losses}.
\newblock In \emph{CVPR}, pages 20626--20636, 2023.

\bibitem[Sch\"{o}nberger and Frahm(2016)]{schoenberger2016sfm}
Johannes~Lutz Sch\"{o}nberger and Jan-Michael Frahm.
\newblock {Structure-from-Motion Revisited}.
\newblock In \emph{CVPR}, 2016.

\bibitem[Sturm et~al.(2012)Sturm, Engelhard, Endres, Burgard, and
  Cremers]{sturm2012benchmark}
J{\"u}rgen Sturm, Nikolas Engelhard, Felix Endres, Wolfram Burgard, and Daniel
  Cremers.
\newblock {A Benchmark for the Evaluation of RGB-D SLAM Systems}.
\newblock In \emph{IROS}, pages 573--580. IEEE, 2012.

\bibitem[Sumikura et~al.(2019)Sumikura, Shibuya, and
  Sakurada]{sumikura2019openvslam}
Shinya Sumikura, Mikiya Shibuya, and Ken Sakurada.
\newblock {OpenVSLAM: A Versatile Visual SLAM Framework}.
\newblock In \emph{ACMMM}, pages 2292--2295, 2019.

\bibitem[Sun et~al.(2017)Sun, Lu, and Yu]{sun2017weighted}
Yule Sun, Ang Lu, and Lu Yu.
\newblock {Weighted-to-Spherically-Uniform Quality Evaluation for
  Omnidirectional Video}.
\newblock \emph{SPL}, 24\penalty0 (9):\penalty0 1408--1412, 2017.

\bibitem[Takikawa et~al.(2021)Takikawa, Litalien, Yin, Kreis, Loop,
  Nowrouzezahrai, Jacobson, McGuire, and Fidler]{takikawa2021neural}
Towaki Takikawa, Joey Litalien, Kangxue Yin, Karsten Kreis, Charles Loop, Derek
  Nowrouzezahrai, Alec Jacobson, Morgan McGuire, and Sanja Fidler.
\newblock {Neural Geometric Level of Detail: Real-Time Rendering with Implicit
  3D Shapes}.
\newblock In \emph{CVPR}, pages 11358--11367, 2021.

\bibitem[Tancik et~al.(2022)Tancik, Casser, Yan, Pradhan, Mildenhall,
  Srinivasan, Barron, and Kretzschmar]{tancik2022blocknerf}
Matthew Tancik, Vincent Casser, Xinchen Yan, Sabeek Pradhan, Ben Mildenhall,
  Pratul~P Srinivasan, Jonathan~T Barron, and Henrik Kretzschmar.
\newblock {Block-NeRF: Scalable Large Scene Neural View Synthesis}.
\newblock In \emph{CVPR}, pages 8248--8258, 2022.

\bibitem[Tancik et~al.(2023)Tancik, Weber, Ng, Li, Yi, Wang, Kristoffersen,
  Austin, Salahi, Ahuja, et~al.]{tancik2023nerfstudio}
Matthew Tancik, Ethan Weber, Evonne Ng, Ruilong Li, Brent Yi, Terrance Wang,
  Alexander Kristoffersen, Jake Austin, Kamyar Salahi, Abhik Ahuja, et~al.
\newblock {Nerfstudio: A Modular Framework for Neural Radiance Field
  Development}.
\newblock In \emph{ACM SIGGRAPH Conference Proceedings}, pages 1--12, 2023.

\bibitem[Teed and Deng(2020)]{teed2020raft}
Zachary Teed and Jia Deng.
\newblock {RAFT: Recurrent All-Pairs Field Transforms for Optical Flow}.
\newblock In \emph{ECCV}, 2020.

\bibitem[Tschernezki et~al.(2021)Tschernezki, Larlus, and
  Vedaldi]{tschernezki21neuraldiff}
Vadim Tschernezki, Diane Larlus, and Andrea Vedaldi.
\newblock {{NeuralDiff}: Segmenting {3D} Objects that Move in Egocentric
  Videos}.
\newblock In \emph{3DV}, 2021.

\bibitem[Turki et~al.(2022)Turki, Ramanan, and Satyanarayanan]{Turki_2022_CVPR}
Haithem Turki, Deva Ramanan, and Mahadev Satyanarayanan.
\newblock {Mega-NeRF: Scalable Construction of Large-Scale NeRFs for Virtual
  Fly-Throughs}.
\newblock In \emph{CVPR}, pages 12922--12931, 2022.

\bibitem[Wang et~al.(2023)Wang, Liu, Chen, Liu, Liu, Komura, Theobalt, and
  Wang]{wang2023f2}
Peng Wang, Yuan Liu, Zhaoxi Chen, Lingjie Liu, Ziwei Liu, Taku Komura,
  Christian Theobalt, and Wenping Wang.
\newblock {F2-NeRF: Fast Neural Radiance Field Training with Free Camera
  Trajectories}.
\newblock In \emph{CVPR}, pages 4150--4159, 2023.

\bibitem[Wang et~al.(2021{\natexlab{a}})Wang, Wang, Genova, Srinivasan, Zhou,
  Barron, Martin-Brualla, Snavely, and Funkhouser]{wang2021ibrnet}
Qianqian Wang, Zhicheng Wang, Kyle Genova, Pratul~P Srinivasan, Howard Zhou,
  Jonathan~T Barron, Ricardo Martin-Brualla, Noah Snavely, and Thomas
  Funkhouser.
\newblock {IBRNet: Learning Multi-View Image-Based Rendering}.
\newblock In \emph{CVPR}, pages 4690--4699, 2021{\natexlab{a}}.

\bibitem[Wang et~al.(2004)Wang, Bovik, Sheikh, and Simoncelli]{wang2004ssim}
Zhou Wang, Alan~C Bovik, Hamid~R Sheikh, and Eero~P Simoncelli.
\newblock {Image Quality Assessment: From Error Visibility to Structural
  Similarity}.
\newblock \emph{TIP}, 13\penalty0 (4):\penalty0 600--612, 2004.

\bibitem[Wang et~al.(2021{\natexlab{b}})Wang, Wu, Xie, Chen, and
  Prisacariu]{wang2021nerf}
Zirui Wang, Shangzhe Wu, Weidi Xie, Min Chen, and Victor~Adrian Prisacariu.
\newblock {NeRF--: Neural Radiance Fields without Known Camera Parameters}.
\newblock \emph{arXiv preprint arXiv:2102.07064}, 2021{\natexlab{b}}.

\bibitem[Weder et~al.(2023)Weder, Garcia-Hernando, Monszpart, Pollefeys,
  Brostow, Firman, and Vicente]{weder2023removing}
Silvan Weder, Guillermo Garcia-Hernando, Aron Monszpart, Marc Pollefeys,
  Gabriel~J Brostow, Michael Firman, and Sara Vicente.
\newblock {Removing Objects from Neural Radiance Fields}.
\newblock In \emph{CVPR}, pages 16528--16538, 2023.

\bibitem[Won et~al.(2019)Won, Ryu, and Lim]{won2019sweepnet}
Changhee Won, Jongbin Ryu, and Jongwoo Lim.
\newblock {SweepNet: Wide-baseline Omnidirectional Depth Estimation}.
\newblock In \emph{ICRA}, pages 6073--6079. IEEE, 2019.

\bibitem[Wu et~al.(2022)Wu, Zhong, Tagliasacchi, Cole, and
  Oztireli]{wu2022d2nerf}
Tianhao Wu, Fangcheng Zhong, Andrea Tagliasacchi, Forrester Cole, and Cengiz
  Oztireli.
\newblock {D2NeRF: Self-Supervised Decoupling of Dynamic and Static Objects
  from a Monocular Video}.
\newblock \emph{NIPS}, 35:\penalty0 32653--32666, 2022.

\bibitem[Zhang et~al.(2018)Zhang, Isola, Efros, Shechtman, and
  Wang]{zhang2018lpips}
Richard Zhang, Phillip Isola, Alexei~A Efros, Eli Shechtman, and Oliver Wang.
\newblock {The Unreasonable Effectiveness of Deep Features as a Perceptual
  Metric}.
\newblock In \emph{CVPR}, pages 586--595, 2018.

\bibitem[Zhang and Scaramuzza(2018)]{zhang2018tutorial}
Zichao Zhang and Davide Scaramuzza.
\newblock {A Tutorial on Quantitative Trajectory Evaluation for Visual
  (-Inertial) Odometry}.
\newblock In \emph{IROS}, pages 7244--725. IEEE, 2018.

\bibitem[Zhou et~al.(2012)Zhou, Yang, and Yu]{zhou2012moving}
Xiaowei Zhou, Can Yang, and Weichuan Yu.
\newblock {Moving Object Detection by Detecting Contiguous Outliers in the
  Low-Rank Representation}.
\newblock \emph{TPAMI}, 35\penalty0 (3):\penalty0 597--610, 2012.

\bibitem[Zioulis et~al.(2018)Zioulis, Karakottas, Zarpalas, and
  Daras]{zioulis2018omnidepth}
Nikolaos Zioulis, Antonis Karakottas, Dimitrios Zarpalas, and Petros Daras.
\newblock {Omnidepth: Dense Depth Estimation for Indoors Spherical Panoramas}.
\newblock In \emph{ECCV}, pages 448--465, 2018.

\bibitem[Zioulis et~al.(2019)Zioulis, Karakottas, Zarpalas, Alvarez, and
  Daras]{zio19spherical}
Nikolaos Zioulis, Antonis Karakottas, Dimitrios Zarpalas, Federico Alvarez, and
  Petros Daras.
\newblock {Spherical View Synthesis for Self-Supervised 360° Depth
  Estimation}.
\newblock In \emph{3DV}, pages 690--699, 2019.

\bibitem[Kingma and Ba(2015)]{KingBa15}
Diederik Kingma and Jimmy Ba.
\newblock {Adam: A Method for Stochastic Optimization}.
\newblock In \emph{ICLR}, 2015.

\bibitem[Martin-Brualla et~al.(2021)Martin-Brualla, Radwan, Sajjadi, Barron,
  Dosovitskiy, and Duckworth]{martin2021nerf}
Ricardo Martin-Brualla, Noha Radwan, Mehdi~SM Sajjadi, Jonathan~T Barron,
  Alexey Dosovitskiy, and Daniel Duckworth.
\newblock {NeRF in the Wild: Neural Radiance Fields for Unconstrained Photo
  Collections}.
\newblock In \emph{CVPR}, pages 7210--7219, 2021.

\bibitem[Mildenhall et~al.(2022)Mildenhall, Verbin, Srinivasan, Hedman,
  Martin-Brualla, and Barron]{multinerf2022}
Ben Mildenhall, Dor Verbin, Pratul~P. Srinivasan, Peter Hedman, Ricardo
  Martin-Brualla, and Jonathan~T. Barron.
\newblock {MultiNeRF}: {A} {Code} {Release} for {Mip-NeRF} 360, {Ref-NeRF}, and
  {RawNeRF}, 2022.

\bibitem[Yuan et~al.(2021)Yuan, Lv, Schmidt, and Lovegrove]{yuan2021star}
Wentao Yuan, Zhaoyang Lv, Tanner Schmidt, and Steven Lovegrove.
\newblock {STaR: Self-supervised Tracking and Reconstruction of Rigid Objects
  in Motion with Neural Rendering}.
\newblock In \emph{CVPR}, pages 13144--13152, 2021.

\end{thebibliography}
}

\end{document}